\documentclass[10pt,twocolumn,letterpaper]{article}

\usepackage{cvpr}
\usepackage{times}
\usepackage{epsfig}
\usepackage{graphicx}
\usepackage{amsmath}
\usepackage{amssymb}

\usepackage{pifont}
\newcommand{\cmark}{\ding{51}}%
\newcommand{\xmark}{\ding{55}}%

\usepackage[utf8]{inputenc} 
\usepackage[T1]{fontenc}    
\usepackage{booktabs}       
\usepackage{amsfonts}       
\usepackage{nicefrac}       
\usepackage{microtype}      
\usepackage{color}
\usepackage{bigstrut}
\usepackage{hhline}
\usepackage{colortbl}
\usepackage{tabularx}
\usepackage{subcaption}
\usepackage[numbers,sort]{natbib} 
\usepackage{arydshln}
\usepackage{array,multirow}
\usepackage{xfrac}
\usepackage{bm}
\usepackage{soul}
\usepackage[export]{adjustbox}
\usepackage{graphicx}
\usepackage{verbatim}
\usepackage{mwe}
\usepackage[shortcuts]{extdash}
\usepackage[toc,page]{appendix}
\usepackage[ruled,lined,linesnumberedhidden]{algorithm2e}
\usepackage{enumitem}
\usepackage{bbold}
\usepackage[normalem]{ulem}
\usepackage[dvipsnames]{xcolor}
\usepackage{bbm}
\usepackage{ulem}
\usepackage{float}
\usepackage{enumitem} 
\usepackage{pdfpages}

\usepackage[pagebackref=true,breaklinks=true,letterpaper=true,colorlinks,bookmarks=false]{hyperref}

\newcolumntype{L}[1]{>{\raggedright\let\newline\\\arraybackslash\hspace{0pt}}m{#1}}
\newcolumntype{R}[1]{>{\raggedleft\let\newline\\\arraybackslash\hspace{0pt}}m{#1}}
\newcolumntype{C}[1]{>{\centering\let\newline\\\arraybackslash\hspace{0pt}}m{#1}}
\newcommand{\rb}{\rotatebox{90}}%

\cvprfinalcopy 

\def\cvprPaperID{3294} 

\ifcvprfinal\fi
\begin{document}

\title{SFNet: Learning Object-aware Semantic Correspondence}

\author{Junghyup Lee\textsuperscript{1,}\thanks{Equal contribution.~$^\dagger$Corresponding author. \newline \textsuperscript{1}School of Electrical and Electronic Engineering, Yonsei University, Seoul, Korea. \newline
\textsuperscript{2}D\'epartement d'Informatique de l'ENS, ENS, CNRS, PSL Research University, Paris, France.} \quad\quad\quad Dohyung Kim\textsuperscript{1,}\footnotemark[1] \quad\quad\quad Jean Ponce\textsuperscript{2,3} \quad\quad\quad Bumsub Ham\textsuperscript{1,$\dagger$}\vspace*{0.2cm}\\
{\textsuperscript{1}Yonsei University \quad\quad \textsuperscript{2}DI ENS \quad\quad \textsuperscript{3}INRIA}}

\maketitle
\thispagestyle{empty}

\begin{abstract}
We address the problem of semantic correspondence, that is, establishing a dense flow field between images depicting different instances of the same object or scene category. We propose to use images annotated with binary foreground masks and subjected to synthetic geometric deformations to train a convolutional neural network (CNN) for this task. Using these masks as part of the supervisory signal offers a good compromise between semantic flow methods, where the amount of training data is limited by the cost of manually selecting point correspondences, and semantic alignment ones, where the regression of a single global geometric transformation between images may be sensitive to image-specific details such as background clutter. We propose a new CNN architecture, dubbed SFNet, which implements this idea. It leverages a new and differentiable version of the argmax function for end-to-end training, with a loss that combines mask and flow consistency with smoothness terms. Experimental results demonstrate the effectiveness of our approach, which significantly outperforms the state of the art on standard benchmarks.
\end{abstract}


\vspace{-0.2cm}
\section{Introduction}\label{sec:introduction}
\vspace{-0.2cm}
\begin{figure}[t]
\captionsetup{font={small}}
\begin{center}
   \includegraphics[width=\columnwidth]{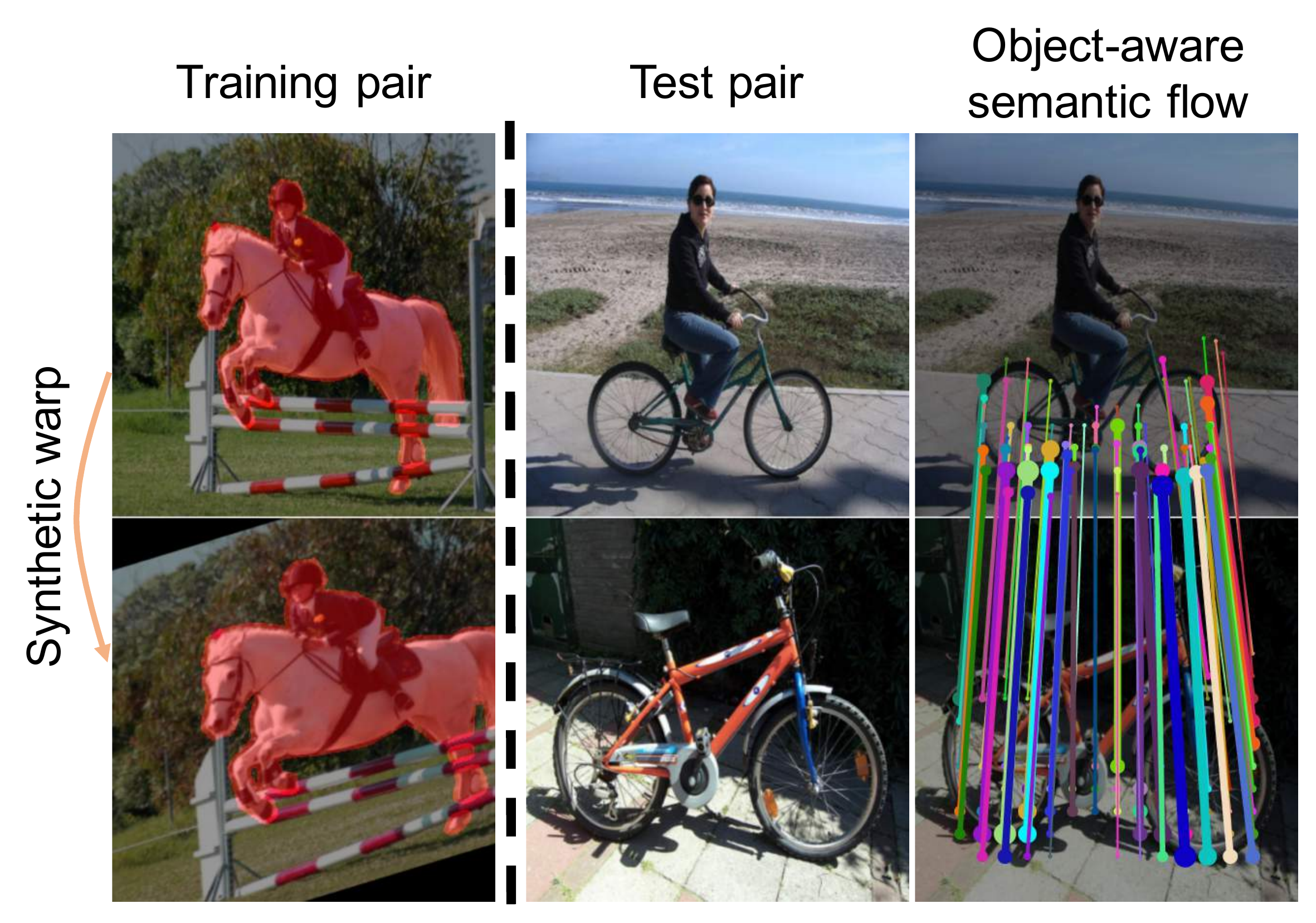}
\end{center}
\vspace{-0.5cm}
   \caption{We use pairs of warped foreground masks obtained from a single image~(left) as a supervisory signal to train our model. This allows us to establish object-aware semantic correspondences across images depicting different instances of the same object or scene category~(right). No masks are required at test time. (Best viewed in color.)}
\vspace{-0.35cm}
\label{fig:teaser}
\end{figure}

Establishing dense correspondences across images is one of the fundamental tasks in computer vision~\cite{okutomi1993multiple,brox2004high,liu2011sift}. Early works have focussed on handling different views of the same scene~(stereo matching~\cite{okutomi1993multiple,hosni2013fast}) or adjacent frames~(optical flow~\cite{brox2004high,brox2009large}) in a video sequence. Semantic correspondence algorithms~(\eg, SIFT Flow~\cite{liu2011sift}) go one step further, finding a dense flow field between images depicting different instances of the same object or scene category. This is very challenging especially in the presence of large changes in appearance/scene layout and background clutter. Classical approaches to semantic correspondence~\cite{liu2011sift,kim2013deformable,hur2015generalized,bristow2015dense,yang2014daisy} typically use an objective function involving fidelity and regularization terms. The fidelity term encourages hand-crafted features~(\eg,~SIFT~\cite{lowe2004distinctive}, HOG~\cite{dalal2005histograms}, DAISY~\cite{tola2010daisy}) to be matched along a dense flow field between images, and the regularization term makes it smooth while aligning discontinuities to object boundaries. Although they have proven useful in various computer vision tasks including object recognition~\cite{liu2011sift,duchenne2011graph}, semantic segmentation~\cite{kim2013deformable}, co-segmentation~\cite{taniai2016joint}, image editing~\cite{dale2009image}, and scene parsing~\cite{kim2013deformable,zhou2015flowweb}, hand-crafted features do not capture high-level semantics~(\eg, appearance and shape variations), and are not robust to image-specific details~(\eg, texture, background clutter, occlusion). 

Convolutional neural networks~(CNNs) have allowed remarkable advances in semantic correspondence in the past few years. Recent methods using CNNs~\cite{han2017scnet,choy2016universal,novotny2017anchornet,kanazawa2016warpnet,kim2017fcss,zhou2016learning,rocco2017convolutional,rocco2018end,Seo2018AttentiveSA,Jeon2018PARN} benefit from rich semantic features invariant to intra-class variations, achieving state-of-the-art results. Semantic flow approaches~\cite{han2017scnet,choy2016universal,novotny2017anchornet,kim2017fcss,zhou2016learning} attempt to find correspondences for individual pixels or patches. They are not seriously affected by non-rigid deformations, but are easily distracted by background clutter. They also require a large amount of data with ground-truth correspondences for training. Although pixel-level semantic correspondences impose very strong constraints, manually annotating them is extremely labor-intensive and somewhat subjective, which limits the amount of training data available~\cite{ham2016proposal}. An alternative is to learn feature descriptor only~\cite{choy2016universal,novotny2017anchornet,kim2017fcss} or to exploit 3D CAD models provided by rendering engines~\cite{zhou2016learning}. 
Semantic alignment methods~\cite{kanazawa2016warpnet,rocco2017convolutional,rocco2018end,Seo2018AttentiveSA,Jeon2018PARN} on the other hand formulate semantic correspondence as a geometric alignment problem and directly regress parameters of a global transformation model~(\eg, affine and thin plate spline) between images. This leverages self-supervised learning where ground-truth parameters are generated synthetically using random transformations with, however, a higher sensitivity to non-rigid deformations. Moreover, background clutter prevents focussing on individual objects and distracts estimating the transformation parameters. To overcome this problem, recent methods alleviate the influence of distractors by inlier counting~\cite{rocco2018end} or an attention process~\cite{Seo2018AttentiveSA}.

In this paper, we present a new approach to establishing an object-aware semantic flow and propose to exploit binary foreground masks as a supervisory signal~(Fig.~\ref{fig:teaser}). Our approach builds upon the insight that correspondences of high quality between images allow to segment common objects from background. To implement this idea, we introduce a new CNN architecture, dubbed SFNet, that outputs a semantic flow field at a sub-pixel level. We leverage a new and differentiable version of the argmax function, a kernel soft argmax, together with mask/flow consistency and smoothness terms to train SFNet end-to-end, establishing object-aware correspondences while filtering out distracting details. Our approach has the following advantages: First, it is a good compromise between current semantic flow and alignment methods, since masks are available for large dataset, and they give a good set of constraints. Exploiting binary foreground masks \emph{explicitly} for training makes it possible to focus on learning correspondences between prominent objects and scene elements. Note that no masks are required at test time. Second, our method establishes a dense non-parametric flow field~(\ie,~semantic flow), which is more robust to non-rigid deformations than a parametric regression~(\ie,~semantic alignment). Finally, the kernel soft argmax enables training the whole network end-to-end, and hence our approach further benefits from high-level semantics specific to the task of semantic correspondence. 
The main contributions of this paper can be summarized as follows:\vspace{-0.2cm}
\begin{itemize}[leftmargin=*]
\item[$\bullet$] We propose to exploit binary foreground masks directly, that are widely available and can be annotated more easily than the pixel-level ground truth, to learn semantic flow by incorporating them into loss functions.\vspace{-0.2cm}
\item[$\bullet$] We introduce a kernel soft argmax, making it less susceptible to multi-modal distributions while providing a differentiable flow field at a sub-pixel level.     
\vspace{-0.2cm}
\item[$\bullet$] We set a new state of the art on standard benchmarks for semantic correspondence, clearly demonstrating the effectiveness of our approach to exploiting foreground masks. We additionally provide an extensive experimental analysis with ablation studies. 
\end{itemize}
\vspace{-0.2cm}
To encourage comparison and future work, our code and models are available online: \url{https://cvlab-yonsei.github.io/projects/SFNet}.

\begin{figure*}
\centering
\captionsetup{font={small}}
\includegraphics[width=0.9\textwidth]{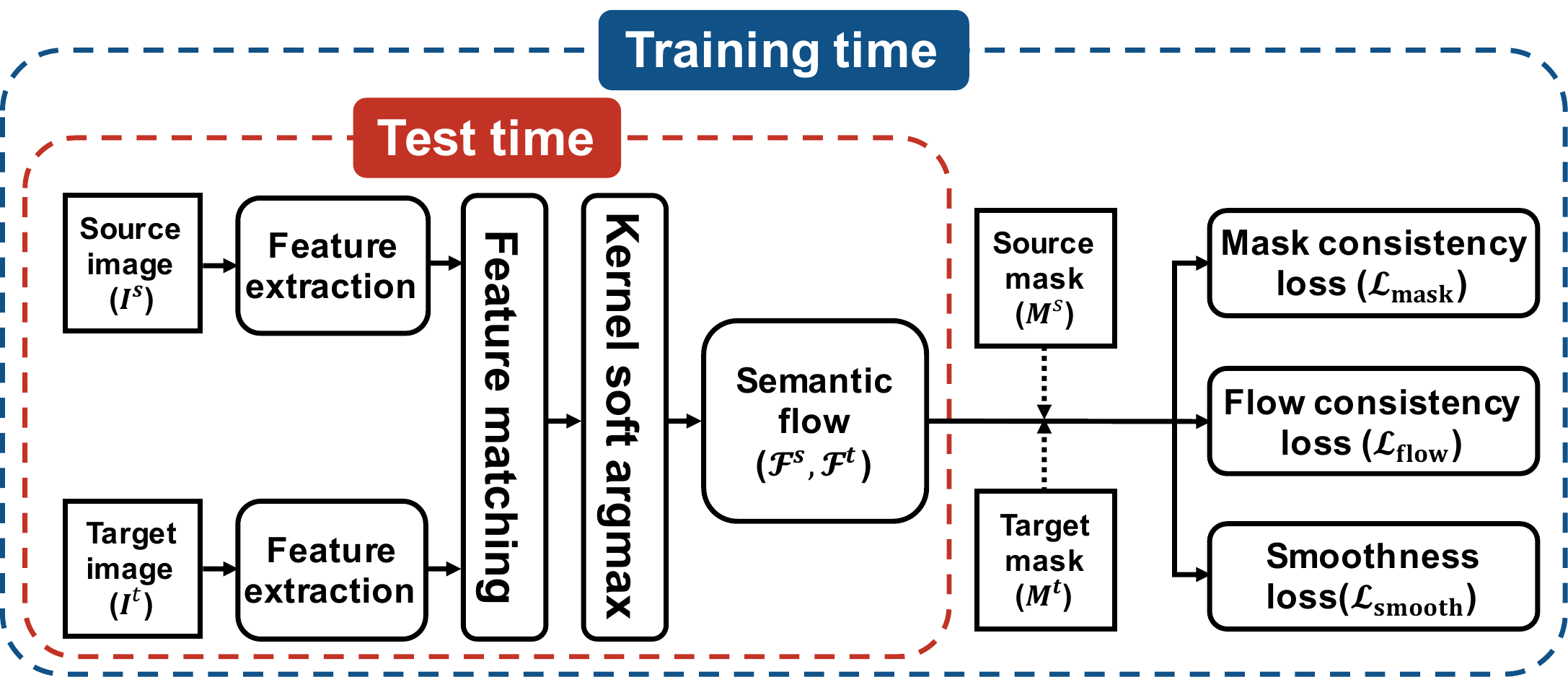} 
\caption{{\bf Overview of SFNet.} SFNet inputs a pair of source and target images, $I^s$~and~$I^t$, and extracts local features using a siamese network. It then computes pairwise matching scores between features and establishes semantic flow,~$\mathcal{F}^{s}$ and $\mathcal{F}^{t}$, for source and target images, respectively, by the kernel soft argmax. At training time, corresponding foreground masks, $M^s$ and $M^t$, for source and target images, respectively, are used to compute mask consistency, flow consistency, and smoothness terms. See text for details.}
\vspace{-0.1cm}
\label{fig:proposed-met}
\end{figure*}

\vspace{-0.2cm}
\section{Related work}
\vspace{-0.2cm}
\label{sec:related}
Correspondence problems cover a broad range of topics in computer vision including stereo, motion analysis, object recognition and shape matching. Giving a comprehensive review on these topics is beyond the scope of this paper. We briefly review representative works related to ours. 

Classical approaches have focussed on finding sparse correspondences,~\eg, for instance matching~\cite{lowe2004distinctive} or establishing dense matches between nearby views of the same scene/object,~\eg,~for stereo matching~\cite{hosni2013fast,okutomi1993multiple} and optical flow estimation~\cite{brox2009large,brox2004high}. Unlike these, semantic correspondence methods estimate dense matches across pictures containing different instances of the same object or scene category. Early works on semantic correspondence focus on matching local features from hand-crafted descriptors, such as SIFT~\cite{liu2011sift,kim2013deformable,hur2015generalized,bristow2015dense}, DAISY~\cite{yang2014daisy} and HOG~\cite{ham2016proposal,taniai2016joint,yang2017object}, together with spatial regularization using graphical models~\cite{liu2011sift,kim2013deformable,taniai2016joint,hur2015generalized} or random sampling~\cite{barnes2009patchmatch,yang2014daisy}.
However, designing hand-crafted features while considering high-level semantics is extremely hard, 
and computing similarities between them is easily distracted~\eg,~by clutter, texture, occlusion and appearance variations. There are many attempts to estimate correspondences robust against background clutter or scale changes between objects/object parts, by using object proposals as candidate regions for matching~\cite{ham2016proposal,yang2017object} or performing matching in scale space~\cite{qiu2014scale}. 

Recently, image features from CNNs have shown the powerful capacity of representing high-level semantics and the robustness to appearance and shape variations~\cite{krizhevsky2012imagenet,simonyan2014very,he2016deep}. Long~\etal~\cite{long2014convnets} first apply CNNs to establish semantic correspondences between images. They follow the same procedure as the SIFT Flow~\cite{liu2011sift} method, but exploit off-the-shelf CNN features trained for ImageNet classification tasks due to a lack of training datasets with pixel-level annotations. This problem can be alleviated by synthesizing ground-truth correspondences from 3D models~\cite{zhou2016learning} or augmenting the number of match pairs in a sparse keypoint dataset using interpolation~\cite{taniai2016joint}. More recently, the PF dataset~\cite{ham2017proposal} has been released providing 1300+ image pairs of 20 image categories with ground-truth annotations from the PASCAL 2011 keypoint dataset~\cite{BourdevMalikICCV09}. This enables learning local features~\cite{han2017scnet,kim2017fcss,novotny2017anchornet} specific to the task of semantic correspondence. Although these approaches using CNN features outperform early methods by large margins, loss functions for training do not involve a spatial regularizer mainly due to a lack of differentiability of the flow field. In contrast, our flow field is differentiable, allowing to train the whole network with a spatial regularizer end-to-end.

Several recent methods~\cite{kanazawa2016warpnet,rocco2017convolutional,rocco2018end,Seo2018AttentiveSA,Jeon2018PARN} formulate semantic correspondence as a geometric alignment problem using parametric models. In particular, these methods first compute feature correlations between images, and they are fed into a regression layer to estimate parameters of a global transformation model~(\eg,~affine, homography, and thin plate spline) to align images. This makes it possible to leverage self-supervised learning~\cite{kanazawa2016warpnet,rocco2017convolutional,rocco2018end,Seo2018AttentiveSA} using synthetically generated data and to train the entire CNNs end-to-end. These approaches apply the same transformation to all pixels, which has the effect of an implicit spatial regularization, providing smooth matches and often outperforming semantic flow methods~\cite{choy2016universal,ham2016proposal,han2017scnet,kim2017fcss,zhou2016learning}. However, they are easily distracted by background clutter and occlusion~\cite{kanazawa2016warpnet,rocco2017convolutional}, since correlations between pairs of features are noisy and include outliers~(\eg,~between different backgrounds). Although this can be alleviated by using attention models~\cite{Seo2018AttentiveSA} or suppressing outlier metches~\cite{rocco2018end}, global transformation models are highly sensitive to non-rigid deformations or local geometric variations. In this context, our method avoids this problem by establishing semantic correspondences directly from feature correlations.
 
Similar to ours, many methods~\cite{Jeon2018PARN,kim2017fcss,zhou2015flowweb,zhou2016learning} leverage object bounding boxes or foreground masks to learn semantic correspondence. They, however, do not incorporate the object location prior explicitly into loss functions. They instead use the prior for pre-processing training samples,~\eg,~generating positive/negative training pairs~\cite{Jeon2018PARN,kim2017fcss} or limiting the candidate regions for matching~\cite{zhou2015flowweb,zhou2016learning}. In contrast, we incorporate the prior directly into loss functions to train the network, outperforming the state of the art by a significant margin.

\vspace{-0.2cm}
\section{Approach}
\vspace{-0.2cm}
In this section, we describe our approach to establishing object-aware semantic correspondences including the network architecture~(Sec.~\ref{sec:archi}) and loss functions~(Sec.~\ref{sec:loss}). An overview of our method is shown in Fig.~\ref{fig:proposed-met}. 

\vspace{-0.2cm}
\subsection{Network architecture}
\vspace{-0.2cm}
\label{sec:archi}
Our model is fully convolutional and mainly consists of three parts~(Fig.~\ref{fig:proposed-met}): We first extract features from source and target images, $I^s$~and~$I^t$, using a siamese network where each sub-network has the same structure with shared parameters. We then compute matching scores between all pairs of local features in the two images, and assign the best match for each feature by the kernel soft argmax. All components are differentiable, allowing us to train the whole network end-to-end. In the following, we describe the network architecture for source to target matching in detail. A target to source matching is similarly computed.

\noindent \textbf{Feature extraction and matching.}
We exploit a ResNet-101~\cite{he2016deep} trained for ImageNet classification~\cite{deng2009imagenet} for feature extraction. Although such CNN features give rich semantics, they typically fire on highly discriminative parts for classification. This may be less adequate for feature matching that requires capturing a spatial deformation for fine-grained localization. We thus use additional adaptation layers to extract features specific to the task of semantic correspondence, transforming them to be highly discriminative w.r.t both appearance and spatial context. This gives a feature map of size $h \times w \times d$ for each image that corresponds to $h \times w$ grids of $d$-dimensional local features. We then apply L2 normalization to the individual $d$-dimensional features. As will be seen in our experiments, the adaptation layers boost the matching performance drastically.

Matching scores are computed using the dot product between local features, resulting in a 4-dimensional correlation map of size~$h \times w \times h \times w$ as follows:
\vspace{-0.1cm}
\begin{equation}
c({\bf{p}},{\bf{q}}) = f^s({\bf{p}})^\top f^t({\bf{q}}),
\vspace{-0.1cm}
\label{eq:norm}
\end{equation}
where we denote by $f^s({\bf{p}})$ and $f^t({\bf{q}})$~$d$-dimensional features at positions ${\bf{p}}=(p_x, p_y)$ and ${\bf{q}}=(q_x, q_y)$ in the source and target images, respectively.

\noindent \textbf{Kernel soft argmax layer.} 
We can assign the best matches by applying the argmax function over a 2-dimensional correlation map~$c_{\bf{p}}({\bf{q}})=c({\bf{p}},{\bf{q}})$,~w.r.t all features~$f^t({\bf{q}})$~at each spatial location~${\bf{p}}$. However, the argmax is discrete and not differentiable. The soft argmax~\cite{honari2018improving,kendall2017end} computes an output by a weighted average of all spatial positions with corresponding matching probabilities. Although it is differentiable and enables fine-grained localization at a sub-pixel level, the output is influenced by all spatial positions, which is problematic especially in the case of multi-modal distributions.

We introduce a hybrid version, the \emph{kernel soft argmax}, that takes advantage of both the soft and discrete argmax. Concretely, it computes correspondences~$\phi({\bf{p}})$ for individual locations~${\bf{p}}$ as an average of all coordinate pairs~${\bf{q}}=(q_x,q_y)$ weighted by a matching probability~$m_{\bf{p}}({\bf{q}})$ as follows. 
\vspace{-0.1cm}
\begin{equation}\label{eq:kernel_soft_max}
	\phi({\bf{p}}) = \sum_{{\bf{q}}}  m_{\bf{p}}({\bf{q}})  {\bf{q}}.
\vspace{-0.1cm}
\end{equation}
The matching probability~$m_{\bf{p}}$ is computed by applying a spatial softmax function to a L2-normalized version~$n_{\bf{p}}$ of the correlation map~$c_{\bf{p}}$: 
\vspace{-0.1cm}
\begin{equation}
	m_{\bf{p}}({\bf{q}}) = \frac{\exp (\beta k_{\bf{p}}({\bf{q}}) n_{\bf{p}}({\bf{q}}))}{\sum_{{\bf{q}}^\prime \in n_{\bf{p}} } \exp (\beta k_{\bf{p}}({\bf{q}}^\prime) n_{\bf{p}}({\bf{q}}^\prime)) },
\vspace{-0.1cm}
\end{equation}
where $k_{\bf{p}}$ is a 2-dimensional Gaussian kernel centered on the position, computed by applying the discrete argmax to~$n_{\bf{p}}$\footnote{At training time, we compute the kernel~$k_{\bf{p}}$ every iterations and no gradients are propagated through the discrete argmax, making the matching probability~$m_{\bf{p}}$ differentiable.}. That is, we perform element-wise multiplication between the score map~$n_{\bf{p}}$ and kernel~$k_{\bf{p}}$, and then apply the softmax function. This retains the scores~$n_{\bf{p}}$ near the output of the discrete argmax while suppressing others, having the effect of restricting the range of averaging in~\eqref{eq:kernel_soft_max} and making it less susceptible to multi-modal distributions~(\eg,~from ambiguous matches in background clutter and repetitive patterns) while maintaining differentiability. $\beta$~is a ``temperature" parameter adjusting a distribution of the softmax output. Note that as it becomes larger, the softmax function approaches the discrete one with one clear peak, but this may cause an unstable gradient flow at training time. Different from~\cite{honari2018improving,kendall2017end}, we perform L2 normalization on the 2-dimensional correlation map~$c_{\bf{p}}$, adjusting the matching scores~$f^s({\bf{p}})^\top f^t({\bf{q}})$ to a common scale before applying the softmax function. Note that the normalization is particularly important for semantic alignment methods~\cite{rocco2017convolutional,rocco2018end,Seo2018AttentiveSA,Jeon2018PARN}~(see, for example, Table~2 in~\cite{rocco2017convolutional}) but for different reasons. It penalizes features having multiple highly-correlated matches, boosting the scores of discriminative matches.

\vspace{-0.2cm}
\subsection{Loss}
\vspace{-0.2cm}
\label{sec:loss}
We exploit binary foreground masks as a supervisory signal to train the network, which gives a strong object prior. To this end, we define three losses that guide the network to learn object-aware correspondences without pixel-level ground truth as
\vspace{-0.1cm}
\begin{equation}
	\mathcal{L} = \lambda_{\mathrm{mask}} \mathcal{L}_{\mathrm{mask}} + 
								   \lambda_{\mathrm{flow}} \mathcal{L}_{\mathrm{flow}} + 
								   \lambda_{\mathrm{smooth}} \mathcal{L}_{\mathrm{smooth}},
   \vspace{-0.1cm}
\end{equation}
which consists of mask consistency~$\mathcal{L}_{\mathrm{mask}}$, flow consistency~$\mathcal{L}_{\mathrm{flow}}$ and smoothness~$\mathcal{L}_{\mathrm{smooth}}$ terms, balanced by the weight parameters~($\lambda_{\mathrm{mask}}$, $\lambda_{\mathrm{flow}}$, $\lambda_{\mathrm{smooth}}$). In the following, we describe each term in detail.  

\noindent \textbf{Mask consistency loss.}
We define a flow field~$\mathcal{F}^s$ from source to target images as 
\vspace{-0.1cm}
\begin{equation}
	\mathcal{F}^{s} ({\bf{p}}) = \phi({\bf{p}}) - {\bf{p}}.
	\vspace{-0.1cm}
\end{equation}
Similarly, a flow field~$\mathcal{F}^t$ from target to source images are defined as~$\phi({\bf{q}}) - {\bf{q}}$. We denote by~$M^s$ and $M^t$ binary masks of source and target images, respectively. The values of 0 and 1 in the masks indicate background and foreground regions, respectively. We assume that reconstructing foreground/background masks by feature matching requires computing reliable similarities between features and dense correspondences of a high quality. To implement this idea, we transfer the target mask~$M^t$ by warping~\cite{jaderberg2015spatial} using the flow field~$\mathcal{F}^s$ and obtain an estimate of the source mask~$\hat M^s$ as follows.
\vspace{-0.1cm}
\begin{equation}
	\hat M^s = \mathcal{W}(M^t ; \mathcal{F}^s).
	\vspace{-0.1cm}
\end{equation}
Here, we denote by~$\mathcal{W}$ a warping operator using a flow field,~\eg,~$\mathcal{W}(M^t ; \mathcal{F}^s)({\bf{p}}) = M^t({\bf{p}} + \mathcal{F}^{s} ({\bf{p}}))$. We then compute the difference between the source mask~$M^s$ and its estimate~$\hat M^s$. Similarly, we reconstruct the target mask~$\hat M^t$ from $M^s$ using the field~$\mathcal{F}^t$ and compute its difference from~$M^t$.  Accordingly, we define the mask consistency loss as
\vspace{-0.1cm}
\begin{equation}
	\mathcal{L}_{\mathrm{mask}} = \sum_{i \in \{s,t \}} \Biggl(\frac{1}{|N^i|} \sum_{{\bf{p}}} ({M}^{i}({\bf{p}}) - \hat {M}^{i}({\bf{p}}))^2\Biggr),
	\vspace{-0.1cm}
\end{equation}
where~$|N^i|$ is the number of pixels in the mask~$M^{i}$. Although the mask consistency loss does not enforce not aligning the background with anything, it prevents matches from foreground to background regions and vice versa by penalizing them. This encourages correspondences to be established between features within foreground masks and background masks, guiding our model to learn object-aware correspondences. Note that the mask consistency loss does not restrict a many-to-one matching. That is, it does not penalize a case when many foreground features in an image are matched to a single one in other image, since binary masks do not give a positional certainty of correspondences. 

\noindent \textbf{Flow consistency loss.}
A flow consistency loss measures consistency between flow fields~$\mathcal{F}^s$ and~$\mathcal{F}^t$ within foreground masks defined as  
\vspace{-0.1cm}
\begin{equation}
	\mathcal{L}_{\mathrm{flow}} = \sum_{i \in \{s,t \}} \Biggl(\frac{1}{|N^i_F|} \sum_{{\bf{p}}} ||(\mathcal{F}^i({\bf{p}}) + \hat{\mathcal{F}}^i({\bf{p}})) \odot M^i({\bf{p}})||^{2}_{2}\Biggr) ,
	\vspace{-0.1cm}
\end{equation}
where~$|N^i_F|$ is the number of foreground pixels in the mask~$M^i$, and
\vspace{-0.1cm}
\begin{equation}\label{eq:flow_warp}
	\hat{\mathcal{F}}^s = \mathcal{W}(\mathcal{F}^t; \mathcal{F}^s),
\vspace{-0.1cm}
\end{equation}
which aligns the flow field~$\mathcal{F}^t$ with respect to~$\mathcal{F}^s$ by warping. $\hat{\mathcal{F}}^t$ is computed similar to~\eqref{eq:flow_warp}. We denote by $\left\lVert \cdot \right\rVert_{2}$ and $\odot$ the L2 norm and element-wise multiplication, respectively. The multiplication is applied separately for each $x$ and $y$ component. The flow consistency term favors a one-to-one matching, spreading flow fields over foreground regions and alleviating the many-to-one matching problem in the mask consistency loss. For example, when the flow fields are consistent with each other,~$\mathcal{F}^s$ and~$\hat{\mathcal{F}}^s$~have the same magnitude with opposite directions. Similar ideas have been explored in stereo matching~\cite{zbontar2015computing,godard2017unsupervised} and optical flow~\cite{meister2018unflow,zou2018df}, but without considering appearance and shape variations. It is hard to incorporate this term in current semantic flow methods based on CNNs~\cite{choy2016universal,han2017scnet,kim2017fcss} mainly due to a lack of differentiability of the flow field. Recently, Zhou~\etal.~\cite{zhou2016learning} exploit cycle consistency between flow fields, but they regress correspondences directly from concatenated features from source and target images and do not consider background clutter. In contrast, our method establishes a differentiable flow field by computing feature similarities explicitly while considering background clutter.

\noindent \textbf{Smoothness loss.}
The differentiable flow field also allows to exploit a smoothness loss, which has been widely used in classical energy-based approaches~\cite{liu2011sift,kim2013deformable,hur2015generalized}. We define a smoothness loss using the first-order derivative of the flow fields~$\mathcal{F}^s$ and~$\mathcal{F}^t$ as
\vspace{-0.15cm}
\begin{equation}
	\mathcal{L}_{\mathrm{smooth}} = \sum_{i \in \{s,t \}} \Biggl(\frac{1}{|N^i_F|} \sum_{{\bf{p}}}
	|| \nabla \mathcal{F}^{i} ({\bf{p}}) \odot M^i({\bf{p}}) ||_1 \Biggr) ,
\end{equation}
where $\left\lVert \cdot \right\rVert_1$ and~$\nabla$ are the L1 norm and the gradient operator, respectively. This regularizes~(or smooths) flow fields within foreground regions while not accounting for correspondences at background. 


\vspace{-0.2cm}
\section{Experiments}
\label{sec:exp}
\vspace{-0.2cm}

In this section we present a detailed analysis and evaluation of our approach including ablation studies on different losses and network architectures. 

\vspace{-0.2cm}
\subsection{Implementation details}
\vspace{-0.2cm}
Following~\cite{rocco2018end,Seo2018AttentiveSA}, we use CNN features from ResNet-101~\cite{he2016deep} trained for ImageNet classification~\cite{deng2009imagenet}. Specifically, we use the networks cropped at $\texttt{conv4-23}$ and $\texttt{conv5-3}$ layers, respectively. This results in two feature maps of size $20 \times 20 \times 1024$ and $10 \times 10 \times 2048$, respectively, for a pair of input images of size~$320 \times 320$, which gives a good compromise between localization accuracy and high-level semantics. Adaptation layers are trained with random initialization, separately for each feature map in a residual fashion~\cite{he2016deep}. To compute residuals, we add $5 \times 5$ and $3\times 3$ convolutional layers with padding on top of $\texttt{conv4-23}$ and $\texttt{conv5-3}$, respectively, with batch normalization~\cite{Ioffe2015BatchNA} and the ReLU~\cite{krizhevsky2012imagenet}. The residuals are then added to the corresponding input features. With the resulting two feature maps of size~$20 \times 20 \times 1024$ and $20 \times 20 \times 2048$\footnote{We upsample the features adapted from $\texttt{conv5-3}$ using bilinear interpolation.}, we compute pairwise match scores and then combine them by element-wise multiplication, resulting in a correlation map of size~$20 \times 20 \times 20 \times 20$. We do not finetune the whole network due to a lack of training data, and train adaptation layers only. We empirically set the temperature parameter~$\beta$ to 50 and standard deviation~$\sigma$ of Gaussian kernel $k_{\bf{p}}$ to 5. Other parameters for losses are fixed to all experiments~($\lambda_{\mathrm{mask}}=3$, $\lambda_{\mathrm{flow}}=16$, $\lambda_{\mathrm{smooth}}=0.5$). We use a grid search to set these parameters, and choose the ones that give the best performance on the validation split of the PF-PASCAL dataset~\cite{ham2017proposal,rocco2018end}. At test time, we upsample a flow field of size $20 \times 20$ using bilinear interpolation.
%

\vspace{-0.2cm}
\subsection{Training} 
\vspace{-0.2cm}
Training our network requires pairs of foreground masks for source and target images depicting different instances of the same object category. Although the TSS~\cite{taniai2016joint} and Caltech-101~\cite{fei2006one} datasets provide such pairs, the number of masks is not enough to train our network~\cite{taniai2016joint} or there is a lack of background clutter~\cite{fei2006one}. Our model trained with these datasets suffers from a overfitting problem or may not generalize well for other images containing clutter. Motivated by~\cite{kanazawa2016warpnet,rocco2017convolutional,Seo2018AttentiveSA,novotny2018self}, we generate pairs of source and target images synthetically from single images by applying random affine transformations and use the synthetically warped pairs as training samples. Corresponding foreground masks are also transformed with the same transformation parameters. Contrary to~\cite{kanazawa2016warpnet,rocco2017convolutional,Seo2018AttentiveSA,novotny2018self}, our model does not perform a parametric regression, and thus it does not require ground-truth transformation parameters for training. We use the Pascal VOC 2012 segmentation dataset~\cite{everingham2010pascal} that consists of 1,464, 1,449, and 1,456 images for training, validation and test, respectively. We exclude 122 images from train/validation sets that overlap with the test split in the PF-PASCAL~\cite{ham2017proposal}, and train our model with the corresponding 2,791 images. We augment the training dataset by horizontal flipping and color jittering. Note that we do not use segmentation masks, provided by the Pascal VOC 2012 dataset, that specify the class of the object at each pixel. We instead generate binary foreground masks using all labeled objects, regardless of image categories and the number of object, at training time. We train our model with a batch size of 16 about 7k iterations, giving roughly 40 epochs over the training data. We use the Adam optimizer~\cite{kingma2014adam} with~$\beta_1 = 0.9$ and~$\beta_2=0.999$. A learning rate initially set to 3e-5 is divided by 5 after 30 epochs. All networks are trained end-to-end using $\texttt{PyTorch}$\cite{paszke2017automatic}.

\setlength{\tabcolsep}{0.3em}
\begin{table}
\small
\begin{center}
\begin{tabular}{@{}C{0.5cm}@{} | @{}C{0.5cm}@{} | L{4.3cm}@{\hspace{1mm}} | @{}C{1.4cm}@{} | @{}C{1.4cm}@{}}

\multicolumn{2}{c|}{\multirow{2}{*}{Type}} & \multicolumn{1}{c|}{\multirow{2}{*}{Methods}} & \multicolumn{2}{c}{PCK ($\alpha=0.1$)}  \\
\multicolumn{2}{c|}{} & \multicolumn{1}{c|}{} & WILLOW    & PASCAL     \\ 
				
\cmidrule{1-5}\morecmidrules\hline
\multirow{5}{*}{\rb{Hand-crafted~}}
&F &DeepFlow~\cite{revaud2016deepmatching} & 0.20 & 0.21\\
&F &GMK~\cite{duchenne2011graph} & 0.27 & 0.27\\
&F &SIFTFlow~\cite{liu2011sift} & 0.38 & 0.33 \\
&F &DSP~\cite{kim2013deformable} & 0.29 & 0.30 \\
&F &HOG+PF-LOM~\cite{ham2017proposal} & 0.56 & 0.45 \\ 
\hline
\multirow{5}{*}{\rb{CNN-based~}}
&A &(T)~ResNet-101+CNNGeo~\cite{rocco2017convolutional} & 0.68 & 0.68\\
&A &(T)~ResNet-101+A2Net~\cite{Seo2018AttentiveSA} & 0.69 & 0.67\\
&A &(T+P)~ResNet-101+WS-SA~\cite{rocco2018end} & \underline{0.71} & \underline{0.72} \\
&F &(B+P)~FCSS+PF-LOM~\cite{kim2017fcss} & 0.58 &0.46 \\
&F &(M)~ResNet-101+Ours & {\bf 0.74} & {\bf 0.79} \\
\hline

\end{tabular}
\captionsetup{font={small}}
\vspace{-2mm}
\caption{Quantitative comparison with the state of the art on the PF-WILLOW~\cite{ham2016proposal} and the test split of the PF-PASCAL~\cite{ham2017proposal,han2017scnet} in terms of the average PCK. We measure the PCK scores with height and width of the bounding box size. All numbers except for the methods of ~\cite{rocco2017convolutional,Seo2018AttentiveSA,rocco2018end} are taken from~\cite{ham2017proposal,Seo2018AttentiveSA}. Numbers in bold indicate the best performance and underscored ones are the second best. We denote by ``F'' and ``A'', respectively, semantic flow and semantic alignment methods. The characters in parentheses are types of a supervisory signal for training;~T: Transformation parameters; P: Image pairs depicting different instances of the same object category; B: Bounding boxes; M: Foreground masks.}

\vspace{-0.3cm}
\label{tab:pf_data_bb}
\end{center}
\end{table}

\vspace{-0.2cm}
\subsection{Results}
\vspace{-0.2cm}
We compare our model to the state of the art on semantic correspondence including hand-crafted and CNN-based methods with the following three benchmark datasets:~PF-WILLOW~\cite{ham2016proposal}, PF-PASCAL~\cite{ham2017proposal}, and  Caltech-101~\cite{fei2006one}. The results for all comparisons have been obtained from the source code or models provided by the authors. 

\noindent \textbf{PF-WILLOW \& PF-PASCAL.}
The PF-WILLOW~\cite{ham2016proposal} and PF-PASCAL~\cite{ham2017proposal} datasets provide 900 and 1,351 image pairs of 4 and 20 image categories, respectively, with corresponding ground-truth object bounding boxes and keypoint annotations. These benchmarks are more challenging than other datasets~\cite{fei2006one,taniai2016joint} for semantic correspondence evaluation, featuring different instances of the same object class in the presence of large changes in appearance and scene layout, clutter and scale changes between objects. To evaluate our model, we use the PF-WILLOW and the test split of the PF-PASCAL provided by~\cite{han2017scnet,rocco2018end} corresponding roughly 900 and 300 image pairs, respectively. We use the probability of correct keypoint~(PCK)~\cite{yang2013articulated} to measure the precision of overall assignment, particularly at sparse keypoints of semantic relevance. We compute the Euclidean distances between warped keypoints using an estimated dense flow and ground truth, and count the number of keypoints whose distances lie within $\alpha \text{max}(h,w)$ pixels, where $\alpha = 0.1$ and $h$ and $w$ are the height and width of the object bounding box, respectively. 

We show in Table~\ref{tab:pf_data_bb} the average PCK scores for the PF-WILLOW and PF-PASCAL datasets, and compare our method with the state of the art including hand-crafted~\cite{revaud2016deepmatching,duchenne2011graph,liu2011sift,kim2013deformable,ham2017proposal} and CNN-based methods~\cite{rocco2017convolutional,Seo2018AttentiveSA,rocco2018end,kim2017fcss}. The PCK scores in~\cite{rocco2017convolutional,Seo2018AttentiveSA,rocco2018end} are obtained by the provided models~(affine + TPS). All other numbers are taken from~\cite{ham2017proposal,Seo2018AttentiveSA}. From this table, we observe four things: (1)~Our model outperforms the state of the art by a significant margin in terms of the PCK especially for the PF-PASCAL datasets. In particular, it shows better performance than other object-aware methods~\cite{ham2017proposal,kim2017fcss} that focus on establishing region correspondences between prominent objects. A plausible explanation is that establishing correspondences between object proposals is susceptible to shape deformations. (2)~We can clearly see that our model gives better results than semantic alignment methods~\cite{rocco2017convolutional,Seo2018AttentiveSA,rocco2018end} on both datasets, but performance gain for the PF-PASCAL dataset, which typically contains pictures depicting a non-rigid deformation and clutter~(\eg, in cat and person classes), is more significant. For example, the PCK gain over WS-SA~\cite{rocco2018end} for the PF-PASCAL~(0.79 vs. 0.72) is about two times more than that for the PF-WILLOW~(0.74 vs. 0.71), indicating that our semantic flow method is more robust to non-rigid deformations and background clutter than semantic alignment approaches. (3)~By comparing our model with a CNN-based semantic flow method~\cite{kim2017fcss}, we can see that involving a spatial regularizer is significant. It focuses on designing fidelity terms~(\eg, using a contrastive loss~\cite{choy2016universal}) only to learn a feature space preserving semantic similarities. This is because of a lack of differentiability of the flow field. In contrast, our model gives a differentiable flow field, allowing to exploit a spatial regularizer while further leveraging high-level semantics from CNN features more specific to semantic correspondence. (4)~We confirm once more a finding in~\cite{long2014convnets} that CNN features trained for ImageNet classification~\cite{deng2009imagenet} clearly show the better ability to handle intra-class variations than hand-crafted ones such as SIFT~\cite{lowe2004distinctive} and HOG~\cite{dalal2005histograms}. 

\begin{figure*}[ht]
\captionsetup{font={small}}
\captionsetup[subfigure]{aboveskip=-0.5pt,belowskip=-0.5pt}
\centering
	\begin{subfigure}{0.16\textwidth} 
	\includegraphics[width=\textwidth, height=0.917\textwidth, frame]{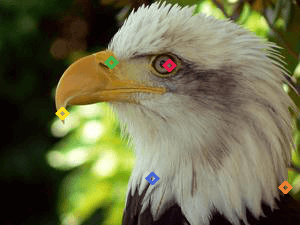}
	\end{subfigure}
	\begin{subfigure}{0.16\textwidth}
	\includegraphics[width=\textwidth, frame]{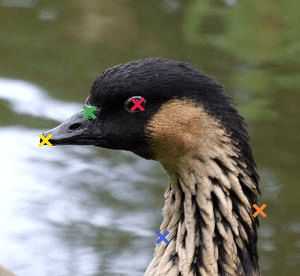}
	\end{subfigure}
	\begin{subfigure}{0.16\textwidth}
	\includegraphics[width=\textwidth, frame]{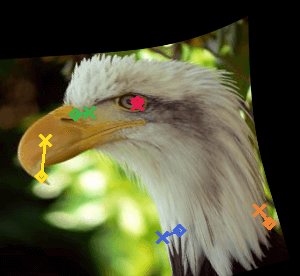}
	\end{subfigure}
	\begin{subfigure}{0.16\textwidth}
	\includegraphics[width=\textwidth, frame]{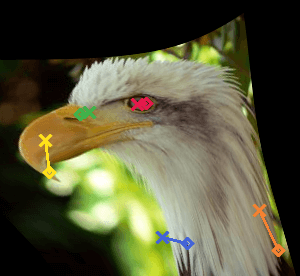}
	\end{subfigure}
	\begin{subfigure}{0.16\textwidth}
	\includegraphics[width=\textwidth, frame]{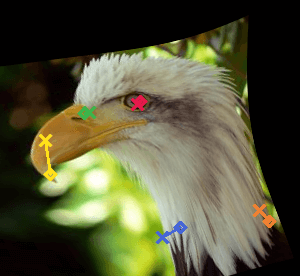}
	\end{subfigure}
	\begin{subfigure}{0.16\textwidth}
	\includegraphics[width=\textwidth, frame]{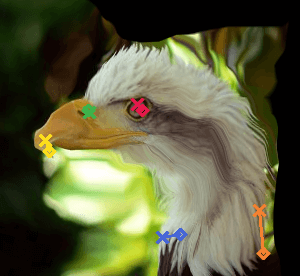}
	\end{subfigure}

	\begin{subfigure}{0.16\textwidth} 
	\includegraphics[width=\textwidth, height=0.685\textwidth, frame]{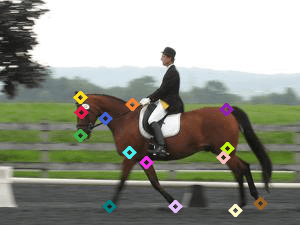}
	\end{subfigure}
	\begin{subfigure}{0.16\textwidth}
	\includegraphics[width=\textwidth, frame]{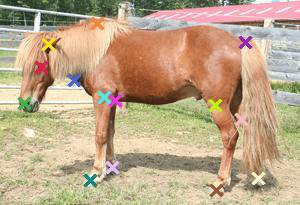}
	\end{subfigure}
	\begin{subfigure}{0.16\textwidth}
	\includegraphics[width=\textwidth, frame]{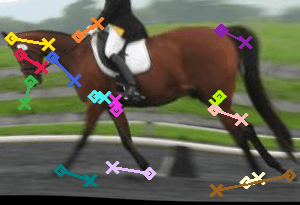}
	\end{subfigure}
	\begin{subfigure}{0.16\textwidth}
	\includegraphics[width=\textwidth, frame]{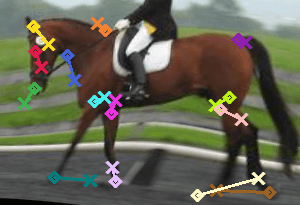}
	\end{subfigure}
	\begin{subfigure}{0.16\textwidth}
	\includegraphics[width=\textwidth, frame]{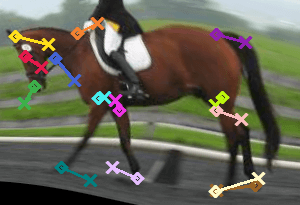}
	\end{subfigure}
	\begin{subfigure}{0.16\textwidth}
	\includegraphics[width=\textwidth, frame]{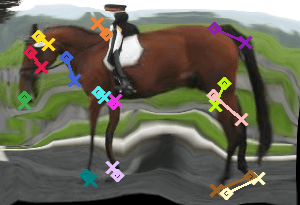}
	\end{subfigure}
	
	\begin{subfigure}{0.16\textwidth} 
	\includegraphics[width=\textwidth, height=1.25\textwidth, frame]{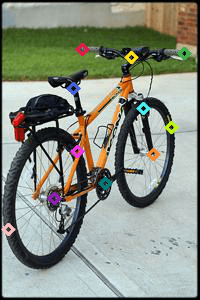}
	\end{subfigure}
	\begin{subfigure}{0.16\textwidth}
	\includegraphics[width=\textwidth, frame]{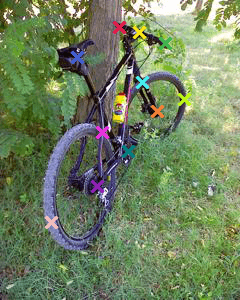}
	\end{subfigure}
	\begin{subfigure}{0.16\textwidth}
	\includegraphics[width=\textwidth, frame]{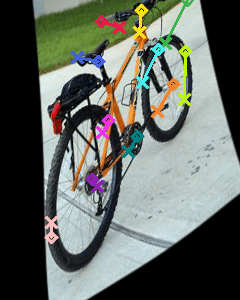}
	\end{subfigure}
	\begin{subfigure}{0.16\textwidth}
	\includegraphics[width=\textwidth, frame]{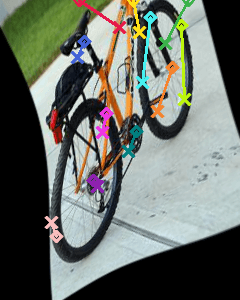}
	\end{subfigure}
	\begin{subfigure}{0.16\textwidth}
	\includegraphics[width=\textwidth, frame]{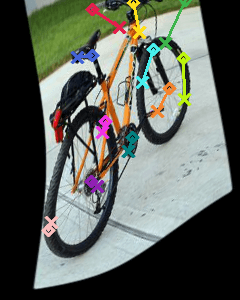}
	\end{subfigure}
	\begin{subfigure}{0.16\textwidth}
	\includegraphics[width=\textwidth, frame]{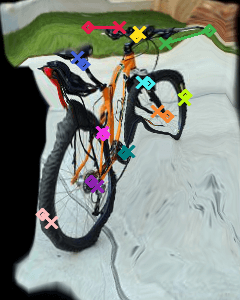}
	\end{subfigure}
	
	\begin{subfigure}{0.16\textwidth} 
	\includegraphics[width=\textwidth, height=0.675\textwidth, frame]{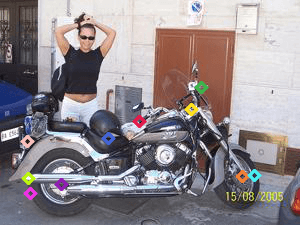}
	\end{subfigure}
	\begin{subfigure}{0.16\textwidth}
	\includegraphics[width=\textwidth, frame]{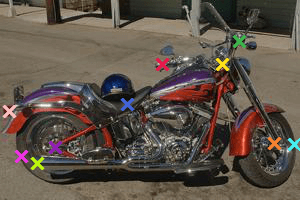}
	\end{subfigure}
	\begin{subfigure}{0.16\textwidth}
	\includegraphics[width=\textwidth, frame]{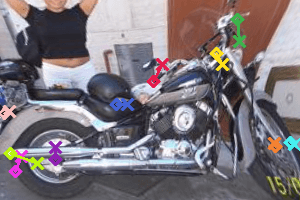}
	\end{subfigure}
	\begin{subfigure}{0.16\textwidth}
	\includegraphics[width=\textwidth, frame]{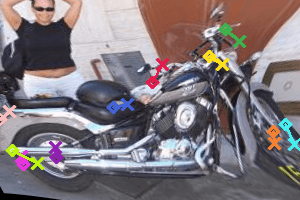}
	\end{subfigure}
	\begin{subfigure}{0.16\textwidth}
	\includegraphics[width=\textwidth, frame]{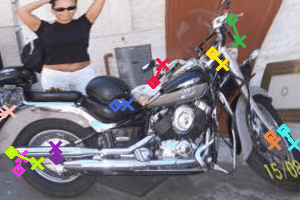}
	\end{subfigure}
	\begin{subfigure}{0.16\textwidth}
	\includegraphics[width=\textwidth, frame]{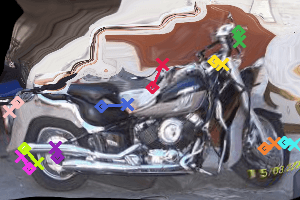}
	\end{subfigure}

	\begin{subfigure}{0.16\textwidth} 
	\includegraphics[width=\textwidth, height=0.75\textwidth, frame]{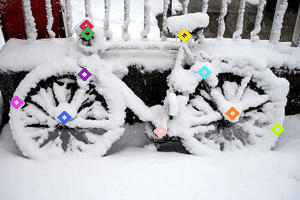}
	\vspace{-0.3cm}
	\caption*{Source image.}
	\end{subfigure}
	\begin{subfigure}{0.16\textwidth}
	\includegraphics[width=\textwidth, frame]{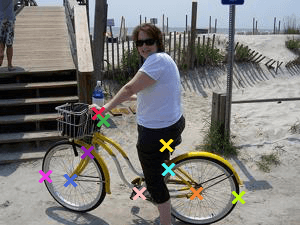}
	\vspace{-0.3cm}
	\caption*{Target image.}
	\end{subfigure}
	\begin{subfigure}{0.16\textwidth}
	\includegraphics[width=\textwidth, frame]{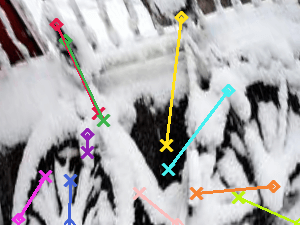}
	\vspace{-0.3cm}
	\caption*{CNNGeo~\cite{rocco2017convolutional}.}
	\end{subfigure}
	\begin{subfigure}{0.16\textwidth}
	\includegraphics[width=\textwidth, frame]{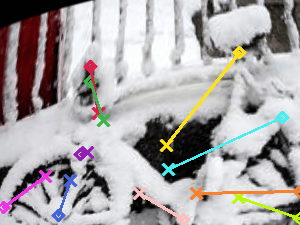}
	\vspace{-0.3cm}
	\caption*{A2Net~\cite{Seo2018AttentiveSA}.}
	\end{subfigure}
	\begin{subfigure}{0.16\textwidth}
	\includegraphics[width=\textwidth, frame]{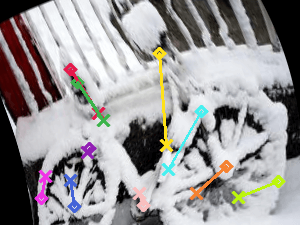}
	\vspace{-0.3cm}
	\caption*{WS-SA~\cite{rocco2018end}.}
	\end{subfigure}
	\begin{subfigure}{0.16\textwidth}
	\includegraphics[width=\textwidth, frame]{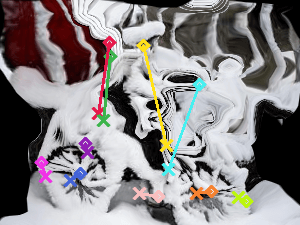}
	\vspace{-0.3cm}
	\caption*{Ours.}
	\end{subfigure}
		
	\vfill
	\vspace{-0.3cm}
\caption{Visual comparison of alignment results between source and target images on the PF-PASCAL dataset~\cite{ham2017proposal}. Keypoints of the source and target images are shown in diamonds and crosses, respectively, with a vector representing the matching error. All methods use the ResNet-101 features. Compared to the state of the art, our method is more robust local non-rigid deformations, scale changes between objects, and clutter. See text for details. (Best viewed in color.)}
\label{fig:qualitative_PF_PASCAL}	
\vspace{-0.3cm}
\end{figure*}
	
\begin{table}[h]
\small
\begin{center}

\begin{tabular}{@{}C{0.5cm}@{} | @{}C{0.5cm}@{} | L{4.3cm}@{\hspace{1mm}} | @{}C{1.4cm}@{} | @{}C{1.4cm}@{}}

\multicolumn{2}{c|}{Type} & \multicolumn{1}{c|}{Methods}  & LT-ACC    & IoU     \\ 
\cmidrule{1-5}\morecmidrules
\hline
\multirow{6}{*}{\rb{Hand-crafted}}
&F &DeepFlow~\cite{revaud2016deepmatching} & 0.74 & 0.40\\
&F &GMK~\cite{duchenne2011graph} & 0.77 & 0.42\\
&F &SIFTFlow~\cite{liu2011sift} & 0.75 & 0.48 \\
&F &DSP~\cite{kim2013deformable} & 0.77 & 0.47 \\
&F&HOG+PF-LOM~\cite{ham2017proposal} & 0.78 & 0.50\\
&F&OADSC~\cite{yang2017object} & 0.81 & 0.55\\
\hline
\multirow{6}{*}{\rb{CNN-based}}
&A&(T)~VGG-16+A2Net~\cite{Seo2018AttentiveSA} & 0.80 & 0.57\\
&A&(T)~ResNet-101+CNNGeo~\cite{rocco2017convolutional} & 0.83 & 0.61\\
&A&(T+P)~ResNet-101+WS-SA~\cite{rocco2018end} & \underline{0.85} & \underline{0.63} \\
&F&(C+P)~VGG-16+SCNet-AG+~\cite{han2017scnet} & 0.79 & 0.51 \\
&F&(B+P)~FCSS+PF-LOM~\cite{kim2017fcss} & 0.83 &0.52 \\
&F&(M)~ResNet-101+Ours & {\bf 0.88} & {\bf 0.67} \\
\hline
\end{tabular}
\captionsetup{font={small}}
\vspace{-2mm}
\caption{Quantitative comparison on the Caltech-101 dataset~\cite{fei2006one}. All numbers are taken from~\cite{ham2017proposal,rocco2018end,Seo2018AttentiveSA}. Numbers in bold indicate the best performance and underscored ones are the second best. C: Ground-truth correspondences.}
\vspace{-0.3cm}
\label{tab:eval_caltech}	
\end{center}
\end{table}

\noindent \textbf{Caltech-101.}
The Caltech-101~\cite{fei2006one} dataset, originally introduced for image classification, provides pictures of 101 image categories with ground-truth object masks. Unlike the PF~\cite{ham2016proposal,ham2017proposal} and TSS~\cite{taniai2016joint} datasets, it does not provide ground-truth keypoint annotations. For fair comparison, we use 15 image pairs, provided by~\cite{han2017scnet,rocco2018end}, for each object category, and use the corresponding 1,515 image pairs for evaluation. Following the experimental protocol in~\cite{kim2013deformable}, we compute matching accuracy with two metrics using the ground-truth masks: Label transfer accuracy~(LT-ACC) and the intersection-over-union~(IoU) metric. Both metrics count the number of correctly labeled pixels between ground-truth and transformed masks using dense correspondences, where the LT-ACC evaluates the overall matching quality while the IoU metric focusses more on foreground objects. Following~\cite{Seo2018AttentiveSA,rocco2018end}, we exclude the LOC-ERR metric, since it measures the localization error of correspondences using object bounding boxes due to a lack of keypoint annotations, which does not cover rotations, affine or deformable transformations. The LT-ACC and IoU comparisons on the Caltech-101 dataset are shown in Table~\ref{tab:eval_caltech}. Although this dataset provides ground-truth object masks, we do not retrain or fine-tune our model to evaluate its generalization ability on other datasets. From this table, we can see that (1)~our model generalizes better than other CNN-based methods for other images outside the training dataset; and~(2) it outperforms the state of the art in terms of the LT-ACC and IoU, verifying once more that our model focuses on regions containing objects while filtering out background clutter, even without using object proposals~\cite{ham2017proposal,han2017scnet,yang2017object,kim2017fcss} or an inlier counting~\cite{rocco2018end}.

%

\begin{table}
\small
\begin{center}
\begin{tabular}{C{1.7cm} C{1.7cm} C{1.7cm} |C{1.3cm}}
				
Mask 			& Flow			 	& \multirow{2}{*}{Smoothness}	& PCK	\\
consistency 	& consistency 		& 				   			 	& ($\alpha=0.1$)  \\ 
\cmidrule{1-4}\morecmidrules\hline
&&& \\[-0.9em]
\cmark		&	\xmark		&	\xmark		&	0.675 				\\
\xmark		&	\cmark		&	\xmark		&	0.718				\\
\cmark		&	\cmark		&	\xmark		&	0.782			 	\\
\cmark		&	\cmark		&	\cmark		&	\textbf{0.787} 		\\
\hline
\end{tabular}
\captionsetup{font={small}}
\vspace{-2mm}
\caption{{Average PCK comparison of different loss functions.}}
\vspace{-0.3cm}
\label{tab:ablation_loss}
\end{center}
\end{table}

\noindent \textbf{Qualitative comparison.}
Figure~\ref{fig:qualitative_PF_PASCAL} shows a visual comparison of alignment results between source and target images with the state of the art on the test split of the PF-PASCAL dataset~\cite{ham2017proposal}. We can see that our method is robust to a local non-rigid deformation~(\eg, bird's beaks and horse's legs in the first two rows), scale changes between objects~(\eg, front wheels in the third row), and clutter~(\eg, wheels in the last row). In particular, the fourth example clearly shows that our method gives more discriminative correspondences, cutting off matches for non-common objects. For example, it does not establish correspondences between a person and background regions in the source and target images, respectively, while others fail to cut off matches on these regions. We can also see that all methods do not handle occlusion~(\eg, a bicycle saddle in the last row).

\vspace{-0.2cm}
\subsection{Ablation study}
\vspace{-0.2cm}
We show an ablation analysis on different components and losses in our model. We measure PCK scores with height and width of the bounding box size, and report the results on the test split of PF-PASCAL dataset~\cite{ham2017proposal,han2017scnet,rocco2018end}. 

\noindent \textbf{Training loss.}
We show the average PCK for three variants of our model in Table~\ref{tab:ablation_loss}. The mask consistency term encourages establishing correspondences between prominent objects. Our model trained with this term only, however, may not yield spatially distinctive correspondences, resulting in the worst performance. A flow consistency term, which spreads flow fields over foreground regions, overcomes this problem, but it does not differentiate correspondences between background and objects. Accordingly, these two terms are complementary each other and exploiting both significantly boosts the performance of our model from 0.675/0.718 to 0.782, already outperforming the state of the art by a large margin~(see Table~\ref{tab:pf_data_bb}). An additional smoothness term further boosts performance to 0.787. 

%
%
%
%

\begin{table}
\small
\begin{center}
\begin{tabular}{C{1.45cm} C{1.45cm} C{0.9cm} C{0.9cm} |C{1.3cm}}
				
Adaptation 	& Multi-level 	& \multicolumn{2}{c|}{Argmax}  			&PCK	\\
layer 		& feature 		& \multicolumn{1}{c|}{Train} & Test     & ($\alpha=0.1$)	 					\\ 
\cmidrule{1-5}\morecmidrules\hline
&&&& \\[-0.9em]
\xmark	&	\cmark	&	-		&	H		&	0.458	 		\\
\xmark	&	\cmark	&	-		&	S		&	0.088    		\\
\xmark	&	\cmark	&	-		&	KS		&	0.284			\\
\hline

\cmark	&	\xmark	&	S		&	H		&	0.725			\\
\cmark	&	\xmark	&	S		&	S		&	0.717			\\
\cmark	&	\xmark	&	KS		&	KS		&	0.750			\\

\hline

\cmark	&	\cmark	&	S		&	H		&	0.768			\\
\cmark	&	\cmark	&	S		&	S		&	0.762			\\
\cmark	&	\cmark	&	KS		&	KS		&	\textbf{0.787}	\\
\hline
\end{tabular}
\captionsetup{font={small}}
\vspace{-2mm}
\caption{Average PCK comparison of different components. We denote by ``H'', ``S'', and ``KS'' hard, soft, and kernel soft argmax operators, respectively.}
\vspace{-0.4cm}
\label{tab:ablation_network}
\end{center}
\end{table}

\noindent \textbf{Network architecture.}
Table~\ref{tab:ablation_network} compares the performance of networks with different components in terms of the average PCK. The baseline models in the first three rows compute matching scores using both features from~$\texttt{conv4-23}$ and $\texttt{conv5-3}$, and estimate correspondences with different argmax operators. They do not involve any training similar to~\cite{long2014convnets} that uses off-the-shelf CNN features for semantic correspondence. We can see that applying the soft argmax directly to the baseline model degrades performance severely, since it is highly susceptible to multi-modal distributions. The results in the next three rows are obtained with a single adaptation layer on top of $\texttt{conv4-23}$. This demonstrates that the adaptation layer extracts features more adequate for pixel-wise semantic correspondences, boosting performance of all baseline models significantly. Particularly, we can see that the kernel soft argmax outperforms others by a large margin, since it enables training our model end-to-end including adaptation layers at a sub-pixel level and is less susceptible to multi-modal distributions. The last three rows suggest that exploiting deeper level of features is important, and using all components with the kernel soft argmax performs best in terms of the average PCK.

\vspace{-0.3cm}
\section{Conclusion}
\vspace{-0.2cm}
We have presented a CNN model for learning an object-aware semantic flow end-to-end, and introduced the corresponding CNN architecture, dubbed SFNet, with a novel kernel soft argmax layer that outputs differential matches at a sub-pixel level. We have proposed to use binary foreground masks directly to train a model for learning pixel-to-pixel correspondences that are widely available and can be obtained easily compared to pixel-level annotations. The ablation studies clearly demonstrate the effectiveness of each component and loss in our model. Finally, we have shown that the proposed method is robust to distracting details and focuses on establishing dense correspondences between prominent objects, outperforming the state of the art on standard benchmarks by a significant margin. 

\noindent \textbf{Acknowledgments.}
This work was supported in part by the National Research Foundation of Korea (NRF) grant funded by the Korea government (MSIP) (No. 2017R1C1B2005584), the Louis Vuitton/ENS chair on artificial intelligence and the NYU/Inria collaboration agreement.

\label{sec:conclusion}


{\small
\bibliographystyle{ieee_fullname}
\bibliography{egbib}

\begin{thebibliography}{10}\itemsep=-1pt

\bibitem{barnes2009patchmatch}
Connelly Barnes, Eli Shechtman, Adam Finkelstein, and Dan~B Goldman.
\newblock {PatchMatch}: A randomized correspondence algorithm for structural
  image editing.
\newblock {\em ACM TOG}, 28(3), 2009.

\bibitem{BourdevMalikICCV09}
Lubomir Bourdev and Jitendra Malik.
\newblock Poselets: Body part detectors trained using 3d human pose
  annotations.
\newblock In {\em ICCV}, 2009.

\bibitem{bristow2015dense}
Hilton Bristow, Jack Valmadre, and Simon Lucey.
\newblock Dense semantic correspondence where every pixel is a classifier.
\newblock In {\em ICCV}, 2015.

\bibitem{brox2009large}
Thomas Brox, Christoph Bregler, and Jitendra Malik.
\newblock Large displacement optical flow.
\newblock In {\em CVPR}, 2009.

\bibitem{brox2004high}
Thomas Brox, Andr{\'e}s Bruhn, Nils Papenberg, and Joachim Weickert.
\newblock High accuracy optical flow estimation based on a theory for warping.
\newblock In {\em ECCV}, 2004.

\bibitem{choy2016universal}
Christopher~B Choy, JunYoung Gwak, Silvio Savarese, and Manmohan Chandraker.
\newblock Universal correspondence network.
\newblock In {\em NIPS}, 2016.

\bibitem{dalal2005histograms}
Navneet Dalal and Bill Triggs.
\newblock Histograms of oriented gradients for human detection.
\newblock In {\em CVPR}, 2005.

\bibitem{dale2009image}
Kevin Dale, Micah~K Johnson, Kalyan Sunkavalli, Wojciech Matusik, and Hanspeter
  Pfister.
\newblock Image restoration using online photo collections.
\newblock In {\em ICCV}, 2009.

\bibitem{deng2009imagenet}
Jia Deng, Wei Dong, Richard Socher, Li-Jia Li, Kai Li, and Li Fei-Fei.
\newblock Image{N}et: {A} large-scale hierarchical image database.
\newblock In {\em CVPR}, 2009.

\bibitem{duchenne2011graph}
Olivier Duchenne, Armand Joulin, and Jean Ponce.
\newblock A graph-matching kernel for object categorization.
\newblock In {\em ICCV}, 2011.

\bibitem{everingham2010pascal}
Mark Everingham, Luc Van~Gool, Christopher~KI Williams, John Winn, and Andrew
  Zisserman.
\newblock {The Pascal Visual Object Classes (VOC) challenge}.
\newblock {\em IJCV}, 88(2), 2010.

\bibitem{fei2006one}
Li Fei-Fei, Rob Fergus, and Pietro Perona.
\newblock One-shot learning of object categories.
\newblock {\em IEEE TPAMI}, 28(4), 2006.

\bibitem{godard2017unsupervised}
Cl{\'e}ment Godard, Oisin Mac~Aodha, and Gabriel~J Brostow.
\newblock Unsupervised monocular depth estimation with left-right consistency.
\newblock In {\em CVPR}, 2017.

\bibitem{ham2016proposal}
Bumsub Ham, Minsu Cho, Cordelia Schmid, and Jean Ponce.
\newblock Proposal flow.
\newblock In {\em CVPR}, 2016.

\bibitem{ham2017proposal}
Bumsub Ham, Minsu Cho, Cordelia Schmid, and Jean Ponce.
\newblock Proposal flow: Semantic correspondences from object proposals.
\newblock {\em IEEE TPAMI}, 40(7), 2018.

\bibitem{han2017scnet}
Kai Han, Rafael~S Rezende, Bumsub Ham, Kwan-Yee~K Wong, Minsu Cho, Cordelia
  Schmid, and Jean Ponce.
\newblock {SCNet}: Learning semantic correspondence.
\newblock In {\em ICCV}, 2017.

\bibitem{he2016deep}
Kaiming He, Xiangyu Zhang, Shaoqing Ren, and Jian Sun.
\newblock Deep residual learning for image recognition.
\newblock In {\em CVPR}, 2016.

\bibitem{honari2018improving}
Sina Honari, Pavlo Molchanov, Stephen Tyree, Pascal Vincent, Christopher Pal,
  and Jan Kautz.
\newblock Improving landmark localization with semi-supervised learning.
\newblock In {\em CVPR}, 2018.

\bibitem{hosni2013fast}
Asmaa Hosni, Christoph Rhemann, Michael Bleyer, Carsten Rother, and Margrit
  Gelautz.
\newblock Fast cost-volume filtering for visual correspondence and beyond.
\newblock {\em IEEE TPAMI}, 35(2), 2013.

\bibitem{hur2015generalized}
Junhwa Hur, Hwasup Lim, Changsoo Park, and Sang Chul~Ahn.
\newblock Generalized deformable spatial pyramid: Geometry-preserving dense
  correspondence estimation.
\newblock In {\em CVPR}, 2015.

\bibitem{Ioffe2015BatchNA}
Sergey Ioffe and Christian Szegedy.
\newblock Batch normalization: Accelerating deep network training by reducing
  internal covariate shift.
\newblock In {\em ICML}, 2015.

\bibitem{jaderberg2015spatial}
Max Jaderberg, Karen Simonyan, Andrew Zisserman, and Koray Kavukcuoglu.
\newblock Spatial transformer networks.
\newblock In {\em NIPS}, 2015.

\bibitem{Jeon2018PARN}
Sangryul Jeon, Seungryong Kim, Dongbo Min, and Kwanghoon Sohn.
\newblock {PARN}: Pyramidal affine regression networks for dense semantic
  correspondence estimation.
\newblock In {\em ECCV}, 2018.

\bibitem{kanazawa2016warpnet}
Angjoo Kanazawa, David~W Jacobs, and Manmohan Chandraker.
\newblock {WarpNet}: Weakly supervised matching for single-view reconstruction.
\newblock In {\em CVPR}, 2016.

\bibitem{kendall2017end}
Alex Kendall, Hayk Martirosyan, Saumitro Dasgupta, Peter Henry, Ryan Kennedy,
  Abraham Bachrach, and Adam Bry.
\newblock End-to-end learning of geometry and context for deep stereo
  regression.
\newblock In {\em ICCV}, 2017.

\bibitem{kim2013deformable}
Jaechul Kim, Ce Liu, Fei Sha, and Kristen Grauman.
\newblock Deformable spatial pyramid matching for fast dense correspondences.
\newblock In {\em CVPR}, 2013.

\bibitem{kim2017fcss}
Seungryong Kim, Dongbo Min, Bumsub Ham, Sangryul Jeon, Stephen Lin, and
  Kwanghoon Sohn.
\newblock {FCSS}: Fully convolutional self-similarity for dense semantic
  correspondence.
\newblock In {\em CVPR}, 2017.

\bibitem{kingma2014adam}
Diederik~P Kingma and Jimmy Ba.
\newblock Adam: A method for stochastic optimization.
\newblock In {\em ICLR}, 2015.

\bibitem{krizhevsky2012imagenet}
Alex Krizhevsky, Ilya Sutskever, and Geoffrey~E Hinton.
\newblock {ImageNet} classification with deep convolutional neural networks.
\newblock In {\em NIPS}, 2012.

\bibitem{liu2011sift}
Ce Liu, Jenny Yuen, and Antonio Torralba.
\newblock {SIFT flow: Dense correspondence across scenes and its applications}.
\newblock {\em IEEE TPAMI}, 33(5), 2011.

\bibitem{long2014convnets}
Jonathan~L Long, Ning Zhang, and Trevor Darrell.
\newblock Do convnets learn correspondence?
\newblock In {\em NIPS}, 2014.

\bibitem{lowe2004distinctive}
David~G Lowe.
\newblock Distinctive image features from scale-invariant keypoints.
\newblock {\em IJCV}, 60(2), 2004.

\bibitem{meister2018unflow}
Simon Meister, Junhwa Hur, and Stefan Roth.
\newblock {UnFlow}: Unsupervised learning of optical flow with a bidirectional
  census loss.
\newblock In {\em AAAI}, 2018.

\bibitem{novotny2018self}
David Novotny, Samuel Albanie, Diane Larlus, and Andrea Vedaldi.
\newblock Self-supervised learning of geometrically stable features through
  probabilistic introspection.
\newblock In {\em CVPR}, 2018.

\bibitem{novotny2017anchornet}
David Novotn{\`y}, Diane Larlus, and Andrea Vedaldi.
\newblock {AnchorNet}: A weakly supervised network to learn geometry-sensitive
  features for semantic matching.
\newblock In {\em CVPR}, 2017.

\bibitem{okutomi1993multiple}
Masatoshi Okutomi and Takeo Kanade.
\newblock A multiple-baseline stereo.
\newblock {\em IEEE TPAMI}, 15(4), 1993.

\bibitem{paszke2017automatic}
Adam Paszke, Sam Gross, Soumith Chintala, Gregory Chanan, Edward Yang, Zachary
  DeVito, Zeming Lin, Alban Desmaison, Luca Antiga, and Adam Lerer.
\newblock Automatic differentiation in {PyTorch}.
\newblock 2017.

\bibitem{qiu2014scale}
Weichao Qiu, Xinggang Wang, Xiang Bai, Alan Yuille, and Zhuowen Tu.
\newblock Scale-space {SIFT F}low.
\newblock In {\em WACV}, 2014.

\bibitem{revaud2016deepmatching}
Jerome Revaud, Philippe Weinzaepfel, Zaid Harchaoui, and Cordelia Schmid.
\newblock {DeepMatching}: Hierarchical deformable dense matching.
\newblock {\em IJCV}, 120(3), 2016.

\bibitem{rocco2017convolutional}
Ignacio Rocco, Relja Arandjelovic, and Josef Sivic.
\newblock Convolutional neural network architecture for geometric matching.
\newblock In {\em CVPR}, 2017.

\bibitem{rocco2018end}
Ignacio Rocco, Relja Arandjelovic, and Josef Sivic.
\newblock End-to-end weakly-supervised semantic alignment.
\newblock In {\em CVPR}, 2018.

\bibitem{Seo2018AttentiveSA}
Paul~Hongsuck Seo, Jongmin Lee, Deunsol Jung, Bohyung Han, and Minsu Cho.
\newblock Attentive semantic alignment with offset-aware correlation kernels.
\newblock In {\em ECCV}, 2018.

\bibitem{simonyan2014very}
Karen Simonyan and Andrew Zisserman.
\newblock Very deep convolutional networks for large-scale image recognition.
\newblock In {\em ICLR}, 2015.

\bibitem{taniai2016joint}
Tatsunori Taniai, Sudipta~N Sinha, and Yoichi Sato.
\newblock Joint recovery of dense correspondence and cosegmentation in two
  images.
\newblock In {\em CVPR}, 2016.

\bibitem{tola2010daisy}
Engin Tola, Vincent Lepetit, and Pascal Fua.
\newblock {DAISY: An efficient dense descriptor applied to wide-baseline
  stereo}.
\newblock {\em IEEE TPAMI}, 32(5), 2010.

\bibitem{yang2017object}
Fan Yang, Xin Li, Hong Cheng, Jianping Li, and Leiting Chen.
\newblock Object-aware dense semantic correspondence.
\newblock In {\em CVPR}, 2017.

\bibitem{yang2014daisy}
Hongsheng Yang, Wen-Yan Lin, and Jiangbo Lu.
\newblock {DAISY filter flow: A generalized discrete approach to dense
  correspondences}.
\newblock In {\em CVPR}, 2014.

\bibitem{yang2013articulated}
Yi Yang and Deva Ramanan.
\newblock Articulated human detection with flexible mixtures of parts.
\newblock {\em IEEE TPAMI}, 35(12), 2013.

\bibitem{zbontar2015computing}
Jure Zbontar and Yann LeCun.
\newblock Computing the stereo matching cost with a convolutional neural
  network.
\newblock In {\em CVPR}, 2015.

\bibitem{zhou2015flowweb}
Tinghui Zhou, Yong Jae~Lee, Stella~X Yu, and Alyosha~A Efros.
\newblock {FlowWeb: Joint image set alignment by weaving consistent, pixel-wise
  correspondences}.
\newblock In {\em CVPR}, 2015.

\bibitem{zhou2016learning}
Tinghui Zhou, Philipp Krahenbuhl, Mathieu Aubry, Qixing Huang, and Alexei~A
  Efros.
\newblock Learning dense correspondence via 3{D}-guided cycle consistency.
\newblock In {\em CVPR}, 2016.

\bibitem{zou2018df}
Yuliang Zou, Zelun Luo, and Jia-Bin Huang.
\newblock {DF-Net}: Unsupervised joint learning of depth and flow using
  cross-task consistency.
\newblock In {\em ECCV}, 2018.

\end{thebibliography}
}
\clearpage


\section*{Supplementary material}

# SFNet: Learning Object-aware Semantic Correspondence
## Supplement

Junghyup Lee[1,*]   Dohyung Kim[1,*]   Jean Ponce[2,3]   Bumsub Ham[1,†]

[1]Yonsei University   [2]DI ENS   [3]INRIA

Here we present a detailed description of the kernel soft argmax and loss functions in Secs. 1 and 2, respectively, and show more quantitative comparisons with the state of the art on three benchmark datasets in Sec. 3: PF-PASCAL [6], PF-WILLOW [5], and TSS [21]. We show alignment examples of a dense flow field on the PF-PASCAL [6], PF-WILLOW [5], TSS [21], and Caltech-101 [4] datasets in Sec. 4. We then discuss issues including training with bounding boxes and with other datasets in Sec. 5.

## 1. Kernel soft argmax

The soft argmax [8, 9] is differentiable, but is susceptible to multi-modal distributions. It computes an output by a weighted average (*i.e.*, an expected value of all spatial coordinates weighted by corresponding probabilities $m_{\mathbf{p}}$). This approximates the discrete argmax only when the matching probability $m_{\mathbf{p}}$ is unimodal with one clear peak. The kernel soft argmax, on the other hand, makes the matching probability $m_{\mathbf{p}}$ have an (approximately) uni-modal distribution, by a 2-dimensional Gaussian kernel $k_{\mathbf{p}}$ centered on the position obtained by the discrete argmax. Note that the center position of the Gaussian kernel is changed every iterations at training time. Note also that the kernel soft argmax is differentiable, since we do not train the Gaussian kernel itself and no gradients are propagated through the discrete argmax. Figure 1 shows an example of estimating correspondences using soft and kernel soft argmax operators. We can clearly see that the soft argmax yields an incorrect correspondence in the presence of multiple highly correlated features, since a weighted average of matching probabilities having multi-modal distributions accumulates positional errors. For example, the soft argmax establishes a correspondence between the points in the background and ship. On the contrary, the Gaussian kernel in the kernel soft argmax suppresses matching probabilities except the ones around the highest value, providing a correct correspondence.

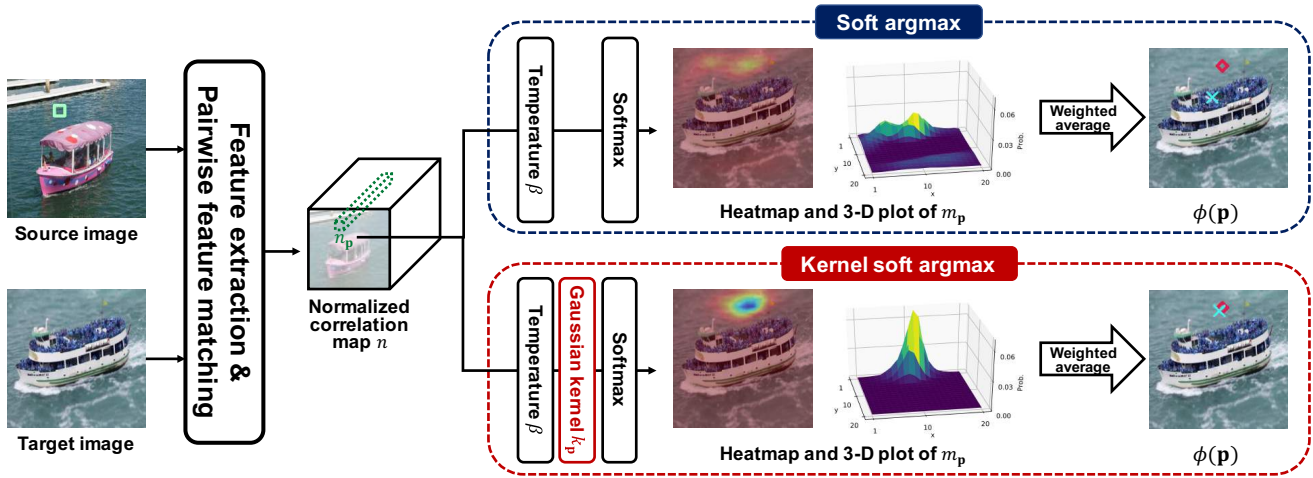

Figure 1: Visualization of soft and kernel soft argmax operations. A point to be matched in the source image is shown in the square. Correspondences computed by either soft or kernel soft argmax operators are shown in crosses. The points, denoted by diamonds, are correspondences established by the discrete argmax. When multiple features are highly correlated, the soft argmax often gives incorrect matches. The kernel soft argmax avoids this problem and approximates the discrete argmax well while maintaining differentiability. (Best viewed in color.)

---
*Equal contribution. †Corresponding author.
[1]School of Electrical and Electronic Engineering, Yonsei University, Seoul, Korea.
[2]Département d'Informatique de l'ENS, ENS, CNRS, PSL Research University, Paris, France.

## 2. Training loss

**Mask and flow consistency losses.** Although a mask consistency term encourages matches between features within foreground/background masks, it may cause a many-to-one matching problem. That is, it does not penalize the case when multiple points are matched to a single one (Fig. 2(a)), since binary masks do not give a positional certainty of correspondences. For example, the foreground mask in the source image can be reconstructed using a single foreground label in the target image. A flow consistency term alleviates this problem. In Fig. 2(b), all points except the center one are penalized by this term. Note that having multiple matches for individual points (*i.e.*, a one-to-many matching) is impossible within our framework. Accordingly, the flow consistency term favors a one-to-one matching (Fig. 2(c)).

**Symmetric loss.** Although the flow consistency loss gives a one-to-one matching, using this loss for a source or a target image only may cause a flow shrinkage problem (Fig. 3(a)). Computing this loss in a symmetric way alleviates this problem. We compute the flow consistency term w.r.t both source and target images, *i.e.*,

$$\mathcal{L}_{\text{flow}} = \sum_{\mathbf{p}} \left( ||(\mathcal{F}^s(\mathbf{p}) + \hat{\mathcal{F}}^s(\mathbf{p})) \odot M^s(\mathbf{p})||_2^2 + ||(\mathcal{F}^t(\mathbf{p}) + \hat{\mathcal{F}}^t(\mathbf{p})) \odot M^t(\mathbf{p})||_2^2 \right). \tag{1}$$

The second term in (1) penalizes the inconsistent matches, *e.g.*, between the entire foreground region in the source image and small parts of the target image (Fig. 3(b)). Using the first or the second term in (1) only does not handle such cases. Note that spreading the flow fields over the entire regions is particularly important to handle scale changes between objects (Fig. 3(c)).

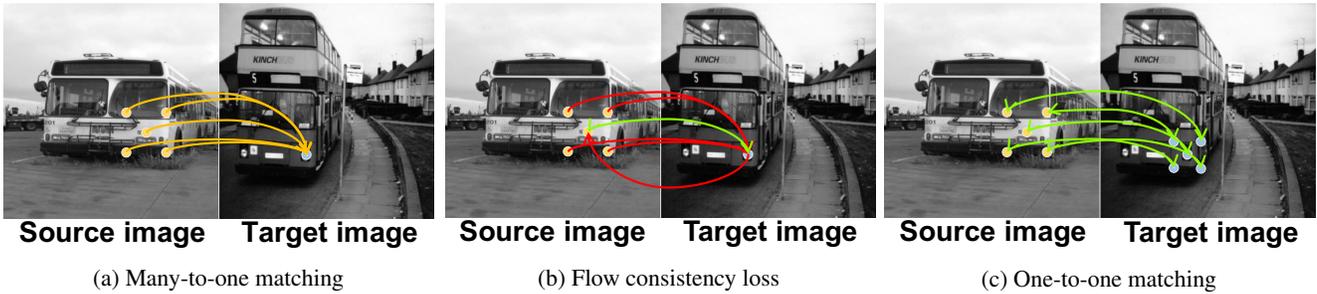

(a) Many-to-one matching  (b) Flow consistency loss  (c) One-to-one matching

Figure 2: Many-to-one matching. (a) Using the mask consistency term only may cause a many-to-one matching problem. Multiple yellow points in the source image can be matched to the single blue one in the target image. The flow consistency term (b) penalizes inconsistent correspondences and (c) favors a one-to-one matching. We denote by green and red arrows consistent and inconsistent matches, respectively.

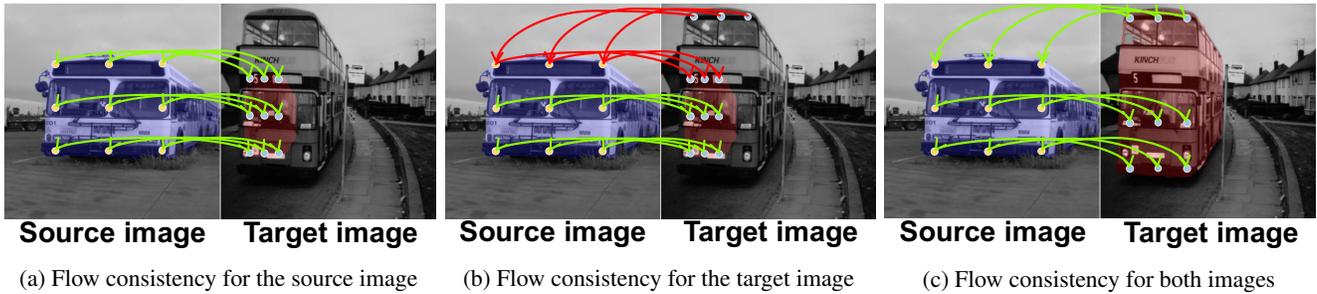

(a) Flow consistency for the source image  (b) Flow consistency for the target image  (c) Flow consistency for both images

Figure 3: Symmetric loss. (a) Considering the flow consistency loss w.r.t a source image may cause a flow shrinkage problem. (b) We penalize inconsistent matches by computing the loss w.r.t a target image as well. (c) This symmetric loss allows to establish an object-to-object matching. We denote by green and red arrows consistent and inconsistent matches, respectively.

## 3. Quantitative results

**PF-WILLOW & PF-PASCAL.** Table 1 shows per-class PCK scores on the PF-PASCAL dataset [6]. We compute the scores in [17, 18, 20] by the provided models (affine + TPS), and take others from [6]. Our model achieves state-of-the-art results for 19 object categories, and outperforms all methods on average by a significant margin. We compare in Table 2 the average PCK scores on the PF datasets. Following [7, 18], they are measured with keypoint coordinates normalized in the range of $[0, 1]$ by dividing them with height and width of the *image size*, respectively, which gives a threshold value larger than the one obtained by the object bounding box in Table 1, resulting in higher scores.

| Method | aero | bike | bird | boat | bot | bus | car | cat | cha | cow | tab | dog | hor | mbik | pers | plnt | she | sofa | trai | tv | **Avg.** |
|---|---|---|---|---|---|---|---|---|---|---|---|---|---|---|---|---|---|---|---|---|---|
| DeepFlow [16] | 0.55 | 0.31 | 0.10 | 0.19 | 0.24 | 0.36 | 0.31 | 0.12 | 0.22 | 0.10 | 0.23 | 0.07 | 0.11 | 0.32 | 0.10 | 0.08 | 0.07 | 0.20 | 0.31 | 0.17 | 0.21 |
| GMK [2] | 0.61 | 0.49 | 0.15 | 0.21 | 0.29 | 0.47 | 0.52 | 0.14 | 0.23 | 0.23 | 0.24 | 0.09 | 0.13 | 0.39 | 0.12 | 0.16 | 0.10 | 0.22 | 0.33 | 0.22 | 0.27 |
| SIFTFlow [15] | 0.61 | 0.56 | 0.20 | 0.34 | 0.32 | 0.54 | 0.56 | 0.26 | 0.29 | 0.21 | 0.33 | 0.17 | 0.23 | 0.43 | 0.18 | 0.17 | 0.17 | 0.31 | 0.41 | 0.34 | 0.33 |
| DSP [10] | 0.64 | 0.56 | 0.17 | 0.27 | 0.37 | 0.51 | 0.55 | 0.29 | 0.23 | 0.24 | 0.19 | 0.15 | 0.23 | 0.41 | 0.15 | 0.11 | 0.18 | 0.27 | 0.35 | 0.27 | 0.39 |
| HOG+PF-LOM [5] | 0.75 | 0.76 | 0.34 | 0.41 | 0.55 | 0.71 | 0.73 | 0.32 | 0.41 | 0.41 | 0.21 | 0.27 | 0.38 | 0.57 | 0.29 | 0.17 | 0.33 | 0.34 | 0.54 | 0.46 | 0.45 |
| ResNet-101+CNNGeo [17] | 0.83 | 0.78 | 0.73 | 0.54 | 0.55 | 0.77 | 0.92 | 0.82 | 0.41 | 0.85 | 0.24 | 0.74 | 0.61 | 0.73 | 0.64 | 0.73 | 0.80 | 0.47 | 0.41 | 0.43 | 0.68 |
| ResNet-101+A2Net [20] | 0.84 | 0.81 | 0.72 | 0.50 | 0.56 | 0.77 | 0.88 | **0.83** | 0.38 | 0.83 | 0.12 | 0.70 | 0.33 | 0.70 | **0.72** | 0.70 | 0.80 | 0.40 | 0.34 | 0.52 | 0.67 |
| ResNet-101+WS-SA [18] | 0.84 | 0.85 | 0.77 | 0.64 | 0.70 | 0.86 | 0.92 | 0.83 | 0.41 | 0.85 | 0.22 | 0.77 | 0.58 | 0.73 | 0.69 | **0.76** | 0.80 | 0.57 | 0.55 | 0.55 | 0.72 |
| Proposed | **0.89** | **0.89** | **0.83** | **0.71** | **0.86** | **0.93** | **0.95** | 0.83 | **0.66** | **0.94** | **0.52** | **0.81** | **0.72** | **0.81** | **0.72** | 0.71 | **1.00** | **0.69** | **0.81** | **0.79** | **0.79** |

Table 1: Per-class PCK on the PF-PASCAL dataset [6].

| Type | | Methods | PCK ($\alpha = 0.1$) | |
|---|---|---|---|---|
| | | | WILLOW | PASCAL |
| CNN-based | F | (C) GoogLeNet+UCN [1] | 0.42 | 0.56 |
| | F | (C) VGG-16+SCNet-AG+ [7] | 0.70 | 0.72 |
| | A | (T) ResNet-101+CNNGeo [17] | 0.81 | 0.72 |
| | A | (T) ResNet-101+A2Net [20] | 0.82 | 0.71 |
| | A | (T+P) ResNet-101+WS-SA [18] | **0.84** | 0.76 |
| | F | (M) ResNet-101+RTNs [11] | - | 0.76 |
| | F | (P) ResNet-101+NCN [19] | - | 0.79 |
| | F | (M) ResNet-101+Ours | **0.84** | **0.82** |

Table 2: Quantitative comparison with the state of the art on the PF-WILLOW [5] and the test split of the PF-PASCAL [6, 7] in terms of the average PCK. We measure the PCK scores with height and width of the image size instead of the bounding box size.

**TSS.** This dataset [21] consists of three subsets (FG3DCar, JODS and PASCAL) that contain 400 image pairs of 7 object categories. It provides dense flow fields obtained by interpolating sparse keypoint matches with additional co-segmentation masks. Following the experimental protocol in [18], we compute the PCK scores densely over the foreground object where the distance threshold is set with $\alpha = 0.05$ and the height and width of the image size. Table 3 compares the average PCK on each subset in the TSS dataset. Our method shows better performance than the state of the art for FG3DCar and JODS. The PASCAL in the TSS contains many image pairs with different poses (*e.g.*, car objects captured with left- and right-side viewpoints). Current methods except for OADSC [22], that is specially designed for handling changes in viewpoints, have a limited capability of finding matches between images with different poses.

| Type | | Methods | FG3D. | JODS | PASC. |
|---|---|---|---|---|---|
| Hand-crafted | F | DSP [10] | 0.487 | 0.465 | 0.382 |
| | F | DFF [23] | 0.493 | 0.303 | 0.224 |
| | F | SIFTFlow [15] | 0.634 | 0.522 | 0.453 |
| | F | HOG+PF-LOM [5] | 0.786 | 0.653 | 0.531 |
| | F | HOG+TSS [21] | 0.830 | 0.595 | 0.483 |
| | F | HOG+OADSC [22] | 0.875 | 0.708 | **0.729** |
| CNN-based | A | (T) VGG-16+A2Net [20] | 0.870 | 0.670 | 0.550 |
| | A | (T) ResNet-101+CNNGeo [17] | 0.901 | 0.764 | 0.563 |
| | A | (T+P) ResNet-101+WS-SA [18] | 0.903 | 0.764 | 0.565 |
| | F | (B+P) FCSS+PF-LOM [12] | 0.839 | 0.635 | 0.582 |
| | F | (B+P) FCSS+DCTM [13] | 0.891 | 0.721 | 0.610 |
| | F | (M) ResNet-101+Ours | **0.906** | **0.787** | 0.565 |

Table 3: Quantitative comparison on the TSS dataset [21]. All numbers are taken from [18, 21].

## 4. Aligned examples

Figures 4, 5, 6, 7 show alignment examples between source and target images on the PF-WILLOW [5], PF-PASCAL [6], TSS [21], and Caltech-101 [4] datasets, respectively. The source images are warped to the target images using dense flow fields established by our model. The alignment examples show that our method establishes semantic correspondences robust to scale changes between objects (*e.g.*, cars in the first two rows in Fig. 4), background clutter (*e.g.*, bikes in the first row in Fig. 5) and local non-rigid deformations (*e.g.*, bird's neck, body, and legs in the fifth and sixth rows in Fig. 7). In Figs. 4, 5 and 6, we also show top 60 matches chosen according to matching probabilities. We can see that most strong matches are established between prominent objects, and matches between foreground and background regions have low matching probabilities. In Fig. 7, we additionally show label transfer results overlaid on target images.

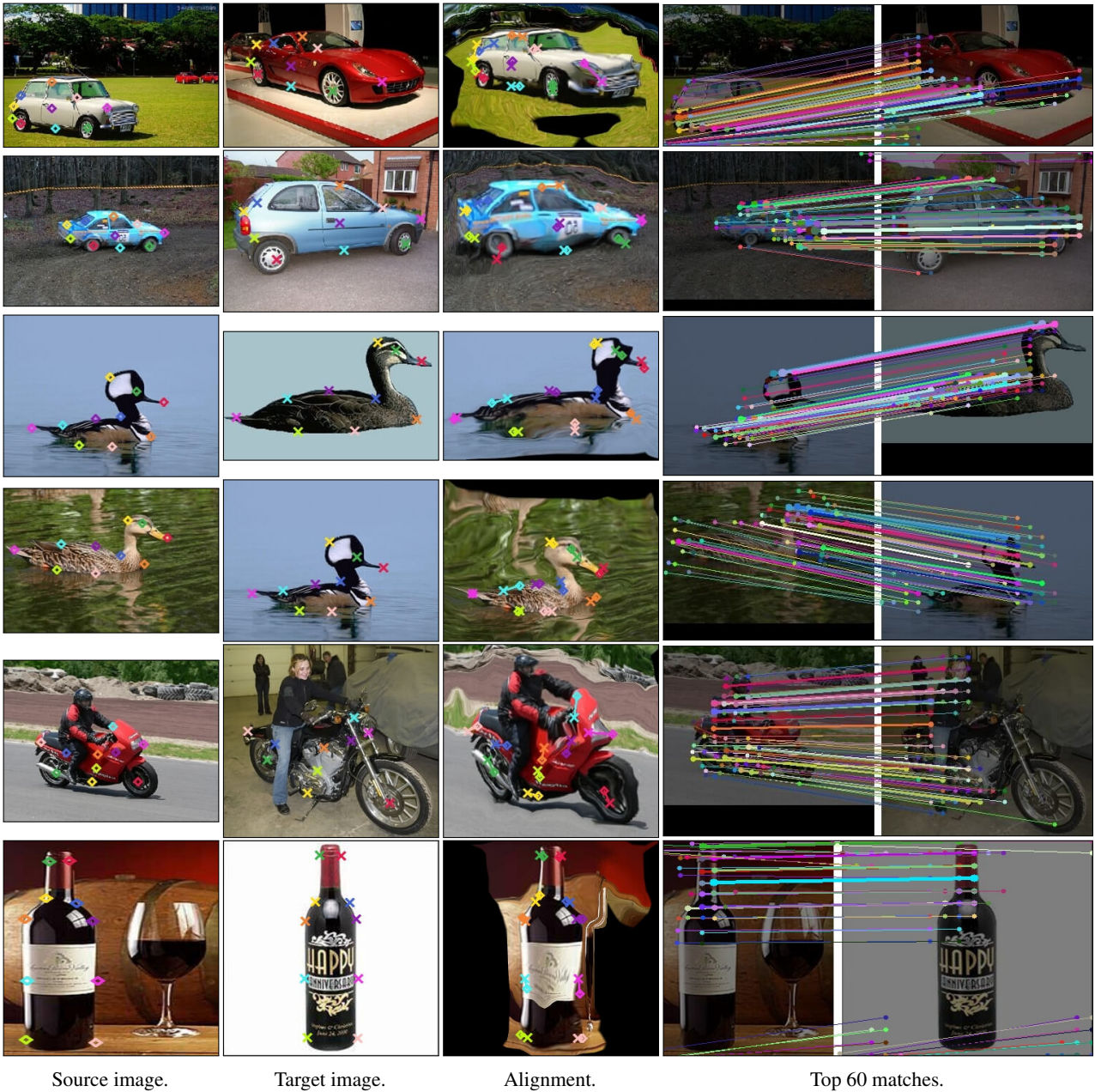

Source image.      Target image.      Alignment.      Top 60 matches.

Figure 4: Alignment examples on the PF-WILLOW dataset [5]. Keypoints of the source and target images are shown in diamonds and crosses, respectively, with a vector representing the matching error. We visualize top 60 matches according to matching probabilities. (Best viewed in color.)

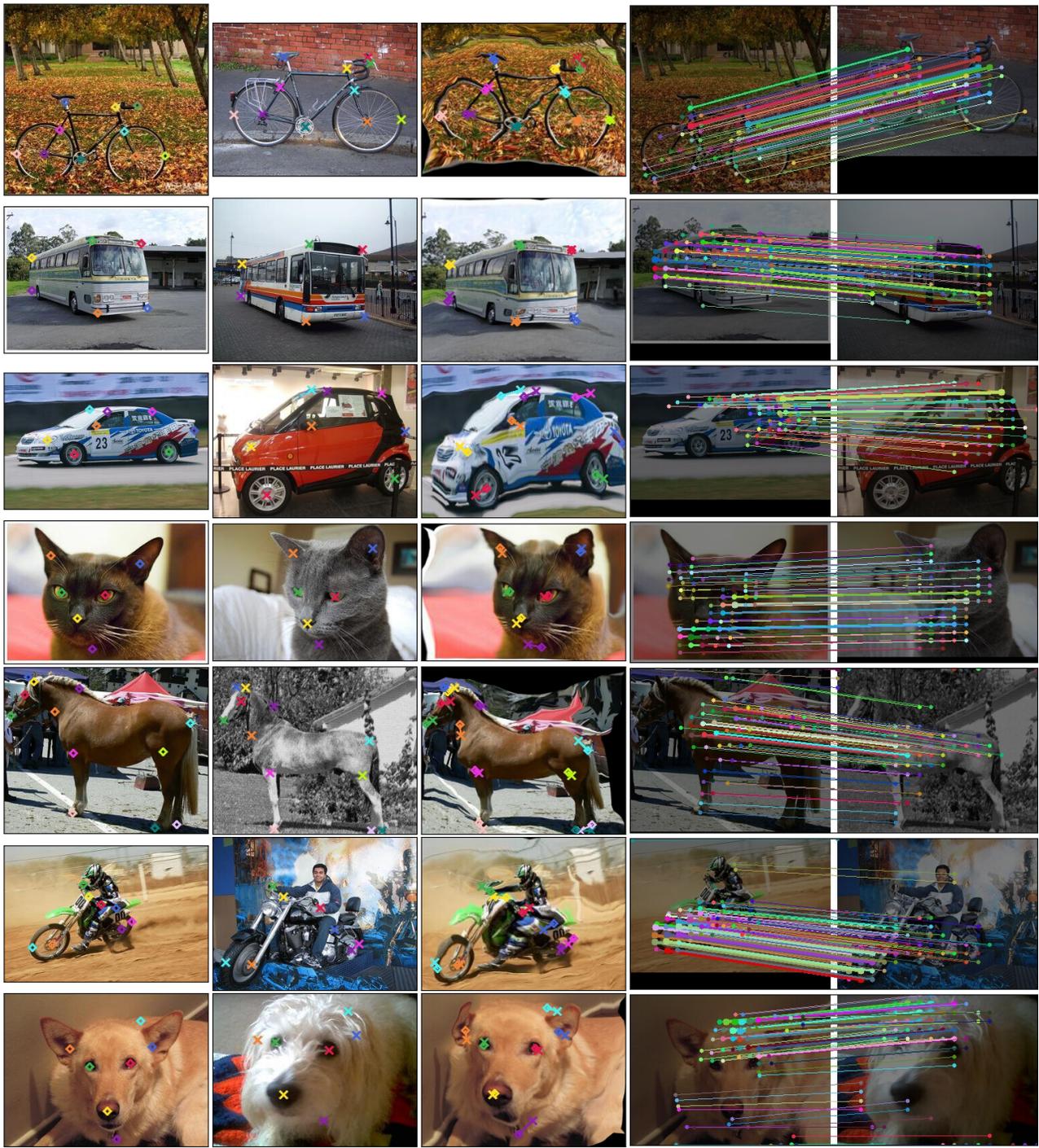

Figure 5: Alignment examples on the PF-PASCAL dataset [6]. Keypoints of the source and target images are shown in diamonds and crosses, respectively, with a vector representing the matching error. We visualize top 60 matches according to matching probabilities. (Best viewed in color.)

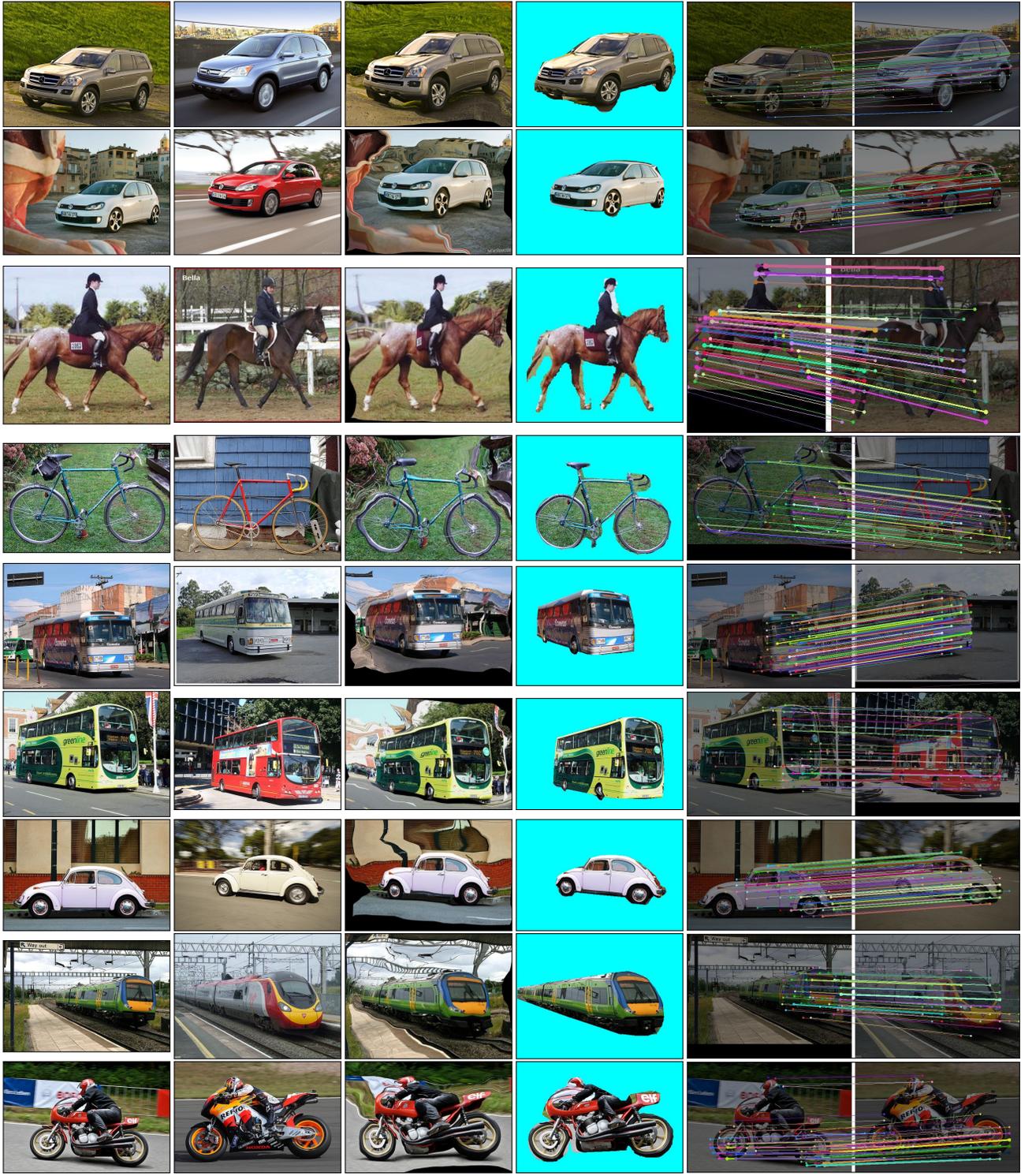

Figure 6: Alignment examples on the TSS dataset [21]. We visualize top 60 matches according to matching probabilities. (Best viewed in color.)

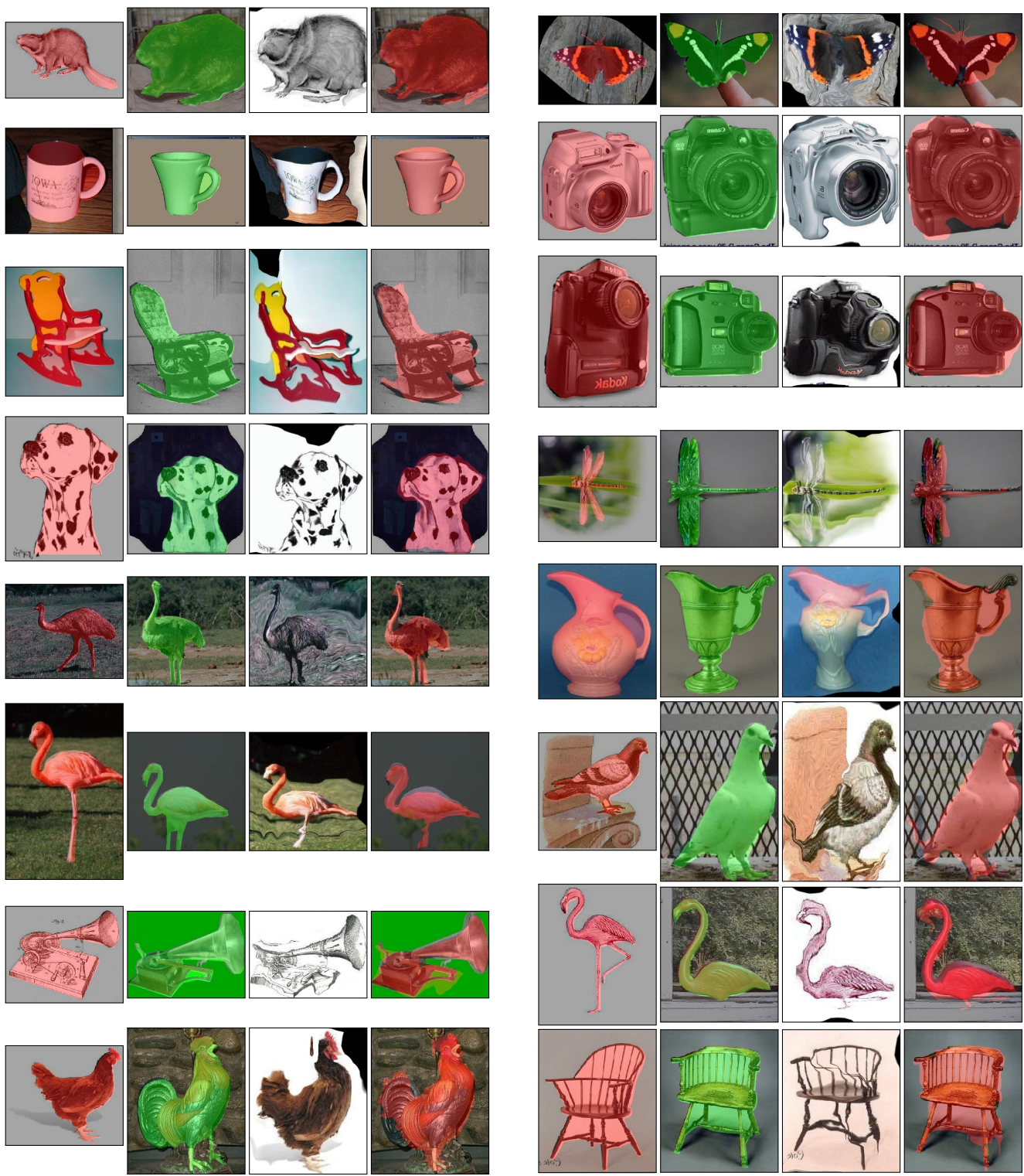

Source image.　　Target image.　　Alignment.　　Label transfer.　　　　Source image.　　Target image.　　Alignment.　　Label transfer.
Figure 7: Alignment examples on the Caltech-101 dataset [4]. The source and target masks are overlaid on corresponding images. We transfer pixel labels of the source images to the target ones using established correspondences. (Best viewed in color.)

## 5. Discussion

**Parameter setting.** We use a grid search with pairs of the temperature parameter $\beta$ and standard deviation $\sigma$ where the maximum search ranges of $\beta$ and $\sigma$ are 100 and 10 with intervals of 10 and 1, respectively. We then select a pair of parameters that gives the best performance on the validation split of the PF-PASCAL dataset [6, 18]. Other parameters ($\lambda_{\text{mask}}$, $\lambda_{\text{flow}}$, $\lambda_{\text{flow}}$) are similarly chosen with the validation split of the PF-PASCAL dataset [6, 18].

**Training with bounding boxes.** We try using the bounding boxes themselves as binary masks during training. The generated masks are noisy, but are less expensive to annotate than ground-truth foreground masks. We use the same 2,791 images from the Pascal VOC 2012 segmentation dataset [3] for training, for an average PCK ($\alpha = 0.1$) of 0.779 on PF-PASCAL [6], which is comparable with the score of 0.787 obtained using ground-truth masks. Using flow consistency terms of (8) w.r.t both source and target images penalizes matches between background and foreground regions, making our method robust to noisy labels.

**Training on PF-PASCAL.** CNN-based methods use different training sets, *e.g.*, the train split of PF-PASCAL [6] (about 700 image pairs) for [7, 18] and Pascal VOC 2011 (11,540 images) for [17, 18, 20]. In [17, 18], the Tokyo Time Machine dataset (20,000 images) is also used. For fair comparison with [7], we train a network on the training split in PF-PASCAL. We excluded 302 images in this split that overlap with either target or source images in the test dataset. Note that [7, 18] ignore this bias. We use object bounding boxes due to the lack of ground-truth foreground masks in the training split. The corresponding average PCK ($\alpha = 0.1$) of 0.778 on PF-PASCAL still outperforms the state of the art (*e.g.*, 0.72 for [18] and 0.68 for [17]) by a large margin.

**Training on larger datasets.** To test this, we use the MS COCO 2014 training dataset [14]. Among 80 object categories, we select 16,624 images of 20 object classes of Pascal VOC 2012 [3] using segmentation masks, which is about $6\times$ the number used in the paper (2,791 images). We test our model on PF-PASCAL [6], since MS COCO does not provide benchmarks for semantic correspondence. Despite using a larger number of training samples, the average PCK ($\alpha = 0.1$) decreases slightly from 0.787 to 0.771, mainly due to domain differences between MS COCO and Pascal VOC. This, however, demonstrates once more the generalization ability of our approach to samples outside the training domain.

## References

[1] Christopher B Choy, JunYoung Gwak, Silvio Savarese, and Manmohan Chandraker. Universal correspondence network. In *NIPS*, 2016. 3
[2] Olivier Duchenne, Armand Joulin, and Jean Ponce. A graph-matching kernel for object categorization. In *ICCV*, 2011. 3
[3] Mark Everingham, Luc Van Gool, Christopher KI Williams, John Winn, and Andrew Zisserman. The Pascal Visual Object Classes (VOC) challenge. *IJCV*, 88(2), 2010. 8
[4] Li Fei-Fei, Rob Fergus, and Pietro Perona. One-shot learning of object categories. *IEEE TPAMI*, 28(4), 2006. 1, 4, 7
[5] Bumsub Ham, Minsu Cho, Cordelia Schmid, and Jean Ponce. Proposal flow. In *CVPR*, 2016. 1, 3, 4
[6] Bumsub Ham, Minsu Cho, Cordelia Schmid, and Jean Ponce. Proposal flow: Semantic correspondences from object proposals. *IEEE TPAMI*, 40(7), 2018. 1, 3, 4, 5, 8
[7] Kai Han, Rafael S Rezende, Bumsub Ham, Kwan-Yee K Wong, Minsu Cho, Cordelia Schmid, and Jean Ponce. SCNet: Learning semantic correspondence. In *ICCV*, 2017. 3, 8
[8] Sina Honari, Pavlo Molchanov, Stephen Tyree, Pascal Vincent, Christopher Pal, and Jan Kautz. Improving landmark localization with semi-supervised learning. In *CVPR*, 2018. 1
[9] Alex Kendall, Hayk Martirosyan, Saumitro Dasgupta, Peter Henry, Ryan Kennedy, Abraham Bachrach, and Adam Bry. End-to-end learning of geometry and context for deep stereo regression. In *ICCV*, 2017. 1
[10] Jaechul Kim, Ce Liu, Fei Sha, and Kristen Grauman. Deformable spatial pyramid matching for fast dense correspondences. In *CVPR*, 2013. 3
[11] Seungryong Kim, Stephen Lin, Sangryul Jeon, Dongbo Min, and Kwanghoon Sohn. Recurrent transformer networks for semantic correspondence. In *NIPS*, 2018.
[12] Seungryong Kim, Dongbo Min, Bumsub Ham, Sangryul Jeon, Stephen Lin, and Kwanghoon Sohn. FCSS: Fully convolutional self-similarity for dense semantic correspondence. In *CVPR*, 2017. 3
[13] Seungryong Kim, Dongbo Min, Stephen Lin, and Kwanghoon Sohn. DCTM: Discrete-continuous transformation matching for semantic flow. In *ICCV*, 2017. 3
[14] Tsung-Yi Lin, Michael Maire, Serge Belongie, James Hays, Pietro Perona, Deva Ramanan, Piotr Dollár, and C Lawrence Zitnick. Microsoft coco: Common objects in context. In *ECCV*, 2014.
[15] Ce Liu, Jenny Yuen, and Antonio Torralba. SIFT flow: Dense correspondence across scenes and its applications. *IEEE TPAMI*, 33(5), 2011. 3

[16] Jerome Revaud, Philippe Weinzaepfel, Zaid Harchaoui, and Cordelia Schmid. DeepMatching: Hierarchical deformable dense matching. *IJCV*, 120(3), 2016. 3
[17] Ignacio Rocco, Relja Arandjelovic, and Josef Sivic. Convolutional neural network architecture for geometric matching. In *CVPR*, 2017. 3, 8
[18] Ignacio Rocco, Relja Arandjelovic, and Josef Sivic. End-to-end weakly-supervised semantic alignment. In *CVPR*, 2018. 3, 8
[19] Ignacio Rocco, Mircea Cimpoi, Relja Arandjelović, Akihiko Torii, Tomas Pajdla, and Josef Sivic. Neighbourhood consensus networks. In *NIPS*, 2018.
[20] Paul Hongsuck Seo, Jongmin Lee, Deunsol Jung, Bohyung Han, and Minsu Cho. Attentive semantic alignment with offset-aware correlation kernels. In *ECCV*, 2018. 3, 8
[21] Tatsunori Taniai, Sudipta N Sinha, and Yoichi Sato. Joint recovery of dense correspondence and cosegmentation in two images. In *CVPR*, 2016. 1, 3, 4, 6
[22] Fan Yang, Xin Li, Hong Cheng, Jianping Li, and Leiting Chen. Object-aware dense semantic correspondence. In *CVPR*, 2017. 3
[23] Hongsheng Yang, Wen-Yan Lin, and Jiangbo Lu. DAISY filter flow: A generalized discrete approach to dense correspondences. In *CVPR*, 2014. 3


\end{document}